\documentclass{article}

\usepackage{adjustbox}
\usepackage[preprint]{neurips_2024}

\newcommand{\bI}{\mathbf{I}}

\newcommand{\bx}{\mathbf{x}}
\newcommand{\by}{\mathbf{y}}

\newcommand{\PP}{\mathbb{P}}
\newcommand{\EE}{\mathbb{E}}
\newcommand{\RR}{\mathbb{R}}
\newcommand{\NN}{\mathbb{N}}

\newcommand{\Tr}[1]{\operatorname{Trace}(#1)}

\newcommand{\tn}[1]{\begin{tiny{#1}\end{tiny}}}

\newcommand{\bo}{\mathbf{0}}

\usepackage[utf8]{inputenc} 
\usepackage[T1]{fontenc}    
\usepackage{hyperref}       
\usepackage{url}            
\usepackage{booktabs}       
\usepackage{subcaption}     
\usepackage{amsfonts}       
\usepackage{nicefrac}       
\usepackage{microtype}      
\usepackage{xcolor} 
\usepackage{float}
\usepackage{siunitx}
\usepackage{amsmath}
\usepackage{amsthm}
\usepackage{graphicx}
\newtheorem{theorem}{Theorem} 
\newtheorem{lemma}{Lemma} 
 
\newtheorem{corollary}{Corollary} 
\newtheorem{definition}{Definition} 
\newtheorem{remark}{Remark} 
 
\usepackage{algorithm}
\usepackage{algpseudocode}
\usepackage{multicol}
\usepackage{pgfplots}
\pgfplotsset{compat=1.17}

\usepgfplotslibrary{colorbrewer}
\usepackage{tikz}
\usetikzlibrary{positioning}
\usepgfplotslibrary{groupplots}

\DeclareMathOperator*{\argmin}{arg\,min} 
\DeclareMathOperator*{\argmax}{arg\,max} 

\title{Efficient Two-Stage Gaussian Process Regression Via  Automatic Kernel Search and Subsampling}
\author{
  Shifan Zhao \\
  Emory University\\
  \texttt{szhao89@emory.edu} \\
  \And
  Jiaying Lu \\
  Emory University\\
  \texttt{jiaying.lu@emory.edu} \\
  \AND
  Ji Yang (Carl) \\
  Emory University \\
  \texttt{j.carlyang@emory.edu} \\
  \And
  Edmond Chow \\
  Georgia Institute of Technology \\
  \texttt{echow@cc.gatech.edu} \\
  \And
  Yuanzhe Xi \\
  Emory University\\
   \texttt{yuanzhe.xi@emory.edu} \\
}

\begin{document}

\maketitle

\begin{abstract}
Gaussian Process Regression (GPR) is widely used in statistics and machine learning for prediction tasks requiring uncertainty measures. Its efficacy depends on the appropriate specification of the \textit{mean function}, \textit{covariance kernel function}, and associated \textit{hyperparameters}. Severe misspecifications can lead to inaccurate results and problematic consequences, especially in safety-critical applications. However, a systematic approach to handle these misspecifications is lacking in the literature. In this work, we propose a general framework to address these issues.
Firstly, we introduce a flexible two-stage GPR framework that separates mean prediction and uncertainty quantification (UQ) to prevent \textbf{mean misspecification}, which can introduce bias into the model. Secondly, \textbf{kernel function misspecification} is addressed through a novel automatic kernel search algorithm, supported by theoretical analysis, that selects the optimal kernel from a candidate set. Additionally, we propose a subsampling-based warm-start strategy for hyperparameter initialization to improve efficiency and avoid \textbf{hyperparameter misspecification}. 
With much lower computational cost, our subsampling-based strategy can yield competitive or better performance than training exclusively on the full dataset.
Combining all these components, we recommend two GPR methods—exact and scalable—designed to match available computational resources and specific UQ requirements. Extensive evaluation on real-world datasets, including UCI benchmarks and a safety-critical medical case study, demonstrates the robustness and precision of our methods\footnote{Two-stage GP github link: \url{https://anonymous.4open.science/r/two-stage-GP-7906}}.
\end{abstract}

\section{Introduction}
\begin{figure}
    \centering
    \includegraphics[width=.99\textwidth]{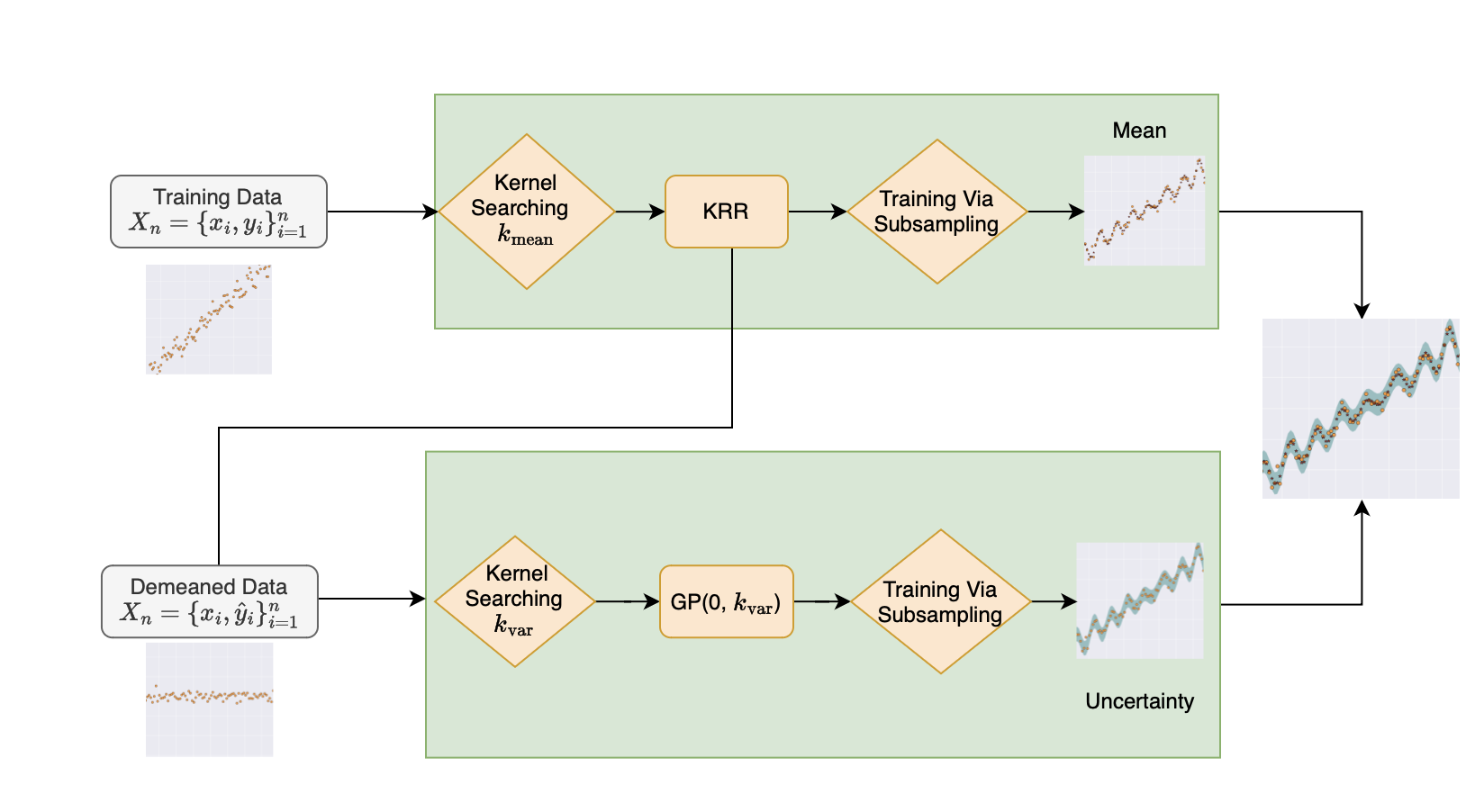}
    \caption{\textbf{Two-stage Gaussian Process Regression (GPR) (Section \ref{sec: mean specification}) via Automatic Kernel Search (Section \ref{sec: kernel misspecification}) and Subsampling (Section \ref{sec: subsampling GP}).}
Stage 1: Automatic Kernel Search selects the best kernel for the mean prediction, followed by mean prediction using a Kernel Ridge Regression (KRR).
Stage 2: After demeaning the training data using the mean prediction from the first stage, automatic Kernel Search identifies the best kernel for uncertainty quantification, and a zero-mean GPR with the corresponding kernel is trained for via subsampling warm start. The final predictive distribution combines these mean and covariance predictions to enhance the model's accuracy and robustness.}
    \label{fig:two stage GP}
\end{figure}

Gaussian processes (GPs) are instrumental in statistics and machine learning, particularly due to their probabilistic predictions, which make them invaluable tools in Bayesian optimization \cite{snoek2012practical}, active learning \cite{rodrigues2014gaussian}, and uncertainty quantification \cite{lu2024uncertainty}. Consider a dataset $X_n = \{\bx_i, y_i\}_{i=1}^{n}$, generated by the relationship $y_i = f(\bx_i) + \epsilon_i$, with $\epsilon_i$ drawn from $\mathcal{N}(0, \sigma_{\epsilon}^2)$ and $\bx_i$ belonging to a subset $\Omega$ of $\mathbb{R}^d$. Gaussian Process Regression (GPR) models this function $f$ using a Gaussian Process with a prior $\mathcal{GP}(m(\cdot), k_{\theta}(\cdot, \cdot))$. Here, \( m(\cdot) \) represents the mean function, and \( k_{\theta}(\bx_i, \bx_j) = \sigma_f^2 c(\bx_i, \bx_j) + \sigma_{\xi}^2 \delta_{ij} \) defines the covariance function where $\theta$ denotes all the hypereparameters in the covariance function. In this covariance function, \( c \) denotes a base kernel function, such as the RBF or Matérn kernel~\footnote{Details of commonly used kernels are provided in Appendix \ref{subsec:kernels}.}. The term \( \delta_{ij} \) is the Kronecker delta, equal to 1 if \( i = j \) and 0 otherwise. Upon conditioning on the observed data, the posterior predictive distribution for a new input $\bx_*$ is calculated as:
\begin{align}
\label{eq: GPR-predictions}
    \hat{m}(\bx_*) &= m(\bx_*) + K_{*X}(\theta)K_{XX}(\theta)^{-1}(\mathbf{y}_n - m(X)), \\
    \hat{k}(\bx_*, \bx_*) &= k_{\theta}(\bx_*, \bx_*) - K_{*X}(\theta)K_{XX}(\theta)^{-1} K_{X*}(\theta),
\end{align}
where $\by_n = [y_1, y_2, \ldots, y_n]^{\top}$, $m(X)$ is a vector obtained via evaluating $m(\cdot)$ over $X_n$, $K_{*X}(\theta)$ and $K_{XX}(\theta)$ denote the kernel matrices obtained by evaluating the covariance function $k_{\theta}$ over $(\bx_*, X_n)$ and $(X_n, X_n)$, respectively.  When the context is clear, we will also use $m_n$ and $K_n$ to denote $m(X)$ and $K_{XX}(\theta)$ for simplicity. Sometimes, when we want to emphasize the hyperparameters $\theta$, we will use $K_n(\theta)$ instead.
Specifying the appropriate mean function $m(\cdot)$ and covariance function $k_{\theta}(\cdot, \cdot)$, with $\theta$ encompassing hyperparameters such as the lengthscale $l$, output scale $\sigma_f^2$, and likelihood noise $\sigma_{\xi}^2$, is pivotal. Typically, the hyperparameters are optimized by minimizing the Negative Log-Likelihood (\(NLL\)) through methods like cross-validation or gradient descent, given by:
\begin{align}
    \label{eq:scaled-likelihood}
    L(\theta; X_n) = \frac{1}{2n}((\by_{n}-m(X))^\top K_{n}^{-1}(\by_{n}-m(X)) + \log{\det{K_n}} + n\log{2\pi}).
\end{align}
Notice in Equation \eqref{eq:scaled-likelihood}, we scale the likelihood with $\frac{1}{n}$, this is a common practice when training GP for stability \cite{Gardner_Pleiss_Bindel_Weinberger_Wilson_2021}. The Gaussian process model trained via the NLL utilizing the full training dataset will be denoted as Exact-GP throughout this paper. Misspecification of the mean and covariance functions can lead to inaccurate or meaningless predictions \cite{coker2023implications, hwang2023use, zhang2004inconsistent}. Our key contributions include a systematic approach to overcome misspecifications called Two-Stage GPR, as shown in Figure \ref{fig:two stage GP}. Section \ref{sec: mean specification} outlines the framework to mitigate mean misspecifications. Section \ref{sec: kernel misspecification} details an automated kernel search algorithm to address kernel misspecification. Finally, we introduce a subsampling warm start strategy in Section \ref{sec: subsampling GP} for efficient training, avoiding hyperparameter misspecifications.

\section{Mitigating Mean Misspecification via Two-stage GPR}
\label{sec: mean specification}
Although a zero-mean prior is commonly used in GPR for simplicity, it often leads to suboptimal reults due to misspecification. A more effective approach is to employ an informative mean prior, which can be established by fitting a mean model initially.
In this section, we explore the impact of mean misspecification and propose a two-stage GPR algorithm to mitigate this issue.
Consider a Gaussian Process \(\mathcal{GP}(m(\cdot), k(\cdot, \cdot))\). The expectation of the NLL, denoted as \(EL\), is computed by taking expectation of Equation \eqref{eq:scaled-likelihood}:
\(
\mathbb{E}[L(\theta; X_n)] 
= \frac{1}{2n}\big(\operatorname{Tr}(K_n(\theta)^{-1} K_n(\theta_*)) + \log \det K_n(\theta) \big),
\label{eq:expectation of NLL}
\)
where \(K_n(\theta)\) and \(K_n(\theta^*)\) represent the covariance matrices under specified parameter \(\theta\) and ground-truth parameter \(\theta^*\) respectively. The first term in \(EL\) quantifies the data fitting, while the second term quantifies the model complexity \cite{williams2006gaussian}. We have ignored the additive constant in Equation \eqref{eq:scaled-likelihood} for simplicity.
In this section, we will use \(EL\) as the loss function instead of \(NLL\) to simplify the theoretical analysis. However, we will demonstrate in Section~\ref{sec: subsampling GP} that a concentration inequality can ensure that \(EL\) and \(NLL\) are closely aligned when the amount of tranining data is large, allowing us to infer similar probabilistic results for \(NLL\) as for \(EL\). If the mean \(m\) is incorrectly assumed to be zero in Equation \eqref{eq:scaled-likelihood}, the hyperparameters are derived by minimizing the misspecified expectation of \(NLL\) (\(MEL\)) instead of \(EL\):
\(
    MEL = \frac{1}{2n}\big(\operatorname{Tr}(K_{n}(\theta)^{-1}[K_{n}(\theta_{*}) - m(X)m(X)^{\top}]) + \log{\det{K_n}}(\theta)\big). \label{eq:MEL}
\)
 The \(MEL\) underestimates the data fitting loss due to the subtraction of a positive term \(m(X)^{\top}K_{n}(\theta)^{-1}m(X)\). Therefore, GPR trained via \(MEL\) usually tends to introduce bias in the estimation in the sense that it penalizes less on the data fitting loss while penalizing more on the model complexity than it should be and often yields an underfitted model. 
In Section \ref{sec: subsampling GP}, we will show that \(\theta^*\) are local minima for \(EL\) under some conditions. 
However, if \(MEL\) is employed, we prove in the following theorem that minimizing \(MEL\) fails to recover \(\theta^*\).
\begin{theorem}
\label{thm: misspecification of mean}
If the mean function $m(x)$ is not a zero function, minimizing the \(MEL\) will not recover the ground-truth hyperparameters $\theta^*$ if $\theta^*$ is not a stationary point of $m(X)^{\top}K_{n}^{-1}\frac{\partial K_n}{\partial \theta} K_{n}^{-1}m(X)$.
\end{theorem}
The theorem indicates that not only is the mean not accurately recovered, but the uncertainty quantification is also compromised due to the misspecified kernel hyperparameters obtained by minimizing \(MEL\). To maintain a concise and fluid article, all proofs are relegated to the appendices corresponding to each section, specifically from Section \ref{appendix: supplementary Materials for sec2} to Section \ref{sec: supplementary Materials for sec4}. 

\begin{figure}
    \centering
    \includegraphics[width=.28\textwidth]{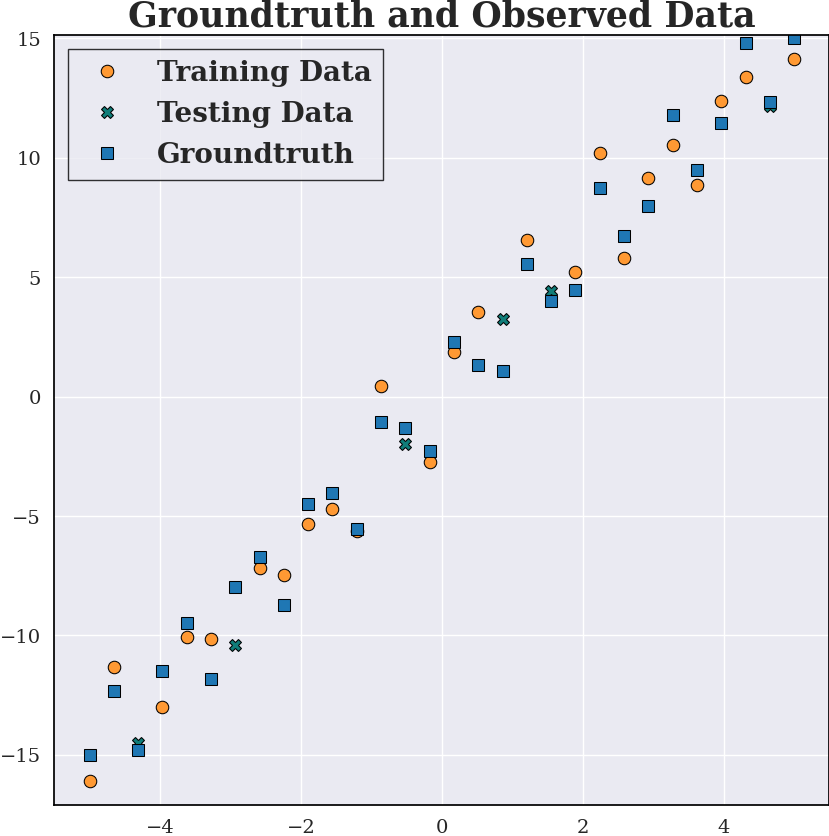}
    \includegraphics[width=.28\textwidth]{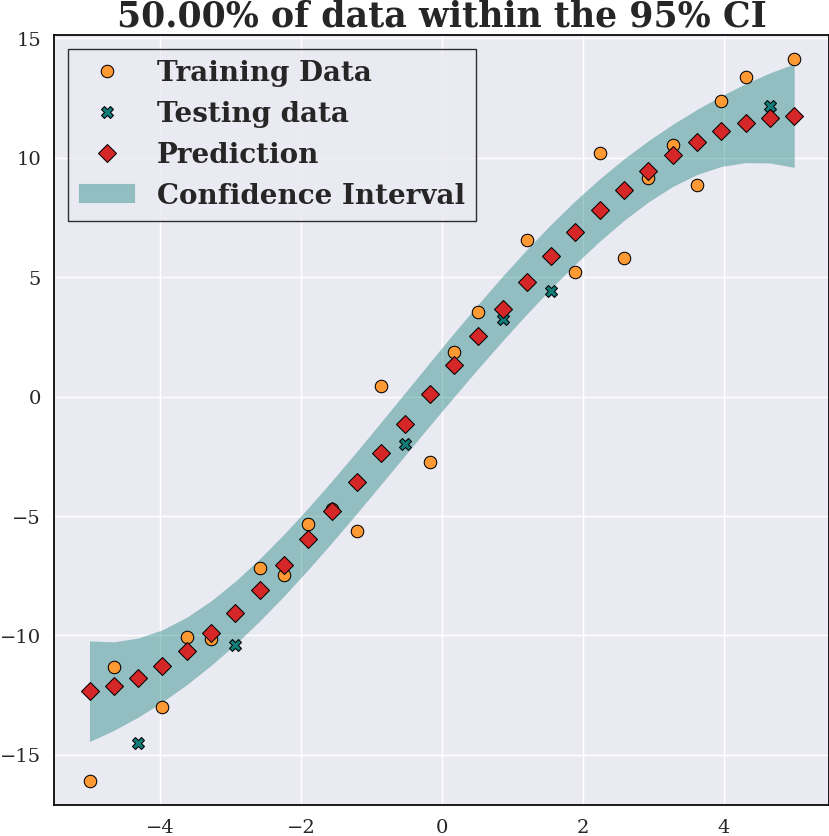}
    \includegraphics[width=.28\textwidth]{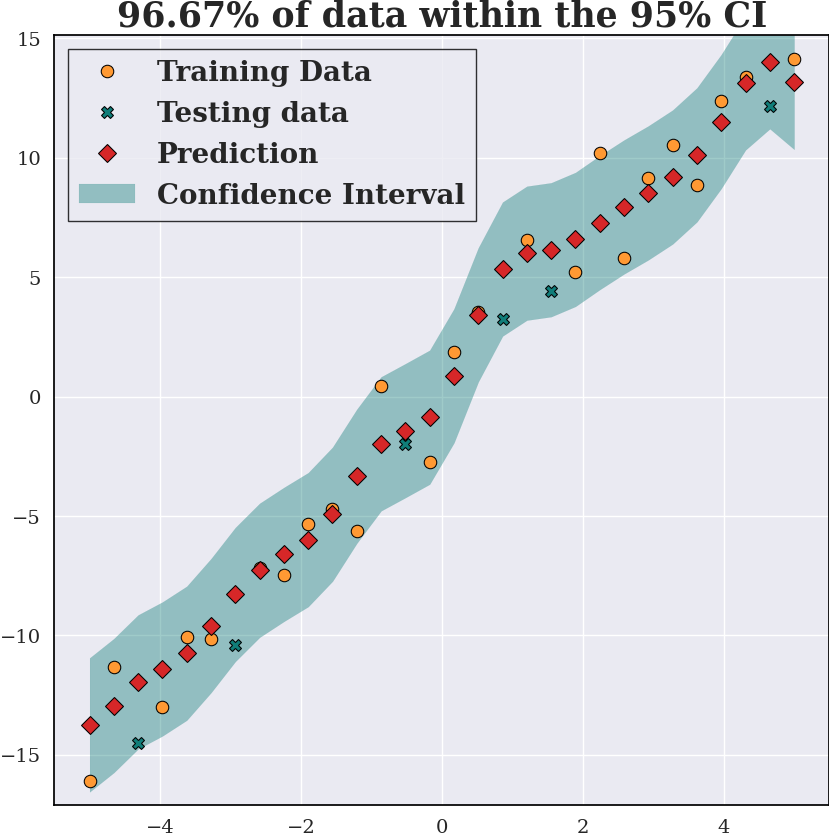}
    \caption{The left panel displays the observed targets $y_i$ and ground-truth function values $f(x_i)$ for 30 points $x_i$ uniformly distributed in the interval $[-5, 5]$. The targets are generated by $y_i = f(x_i) + \epsilon_i$, where $f(x) = 3x + 2\sin(2\pi x)$ and $\epsilon_i \sim N(0, 1)$. The dataset is randomly split into 80\% training and 20\% testing dataset. The middle panel illustrates predictions and 95\% confidence intervals from a single-stage GP using a zero-mean prior. The right panel presents results from a two-stage GP approach. The Exact-GP and two-stage GP are both trained with same optimizer and learning rate. More details can be found in Appendix \ref{appendix: supplementary Materials for sec2}.
    \textbf{Only 50\% of the data are covered by the 95\% confidence interval provided by Exact-GP, whereas our method covers 96.67\% of the data.}}
    \label{fig:proposition-1-few-data-simplefunc}
\end{figure}

To illustrate the impact caused by mean misspecification, we utilize a toy example. 
The leftmost plot in Figure \ref{fig:proposition-1-few-data-simplefunc} displays both the function values and the observed targets. The central plot illustrates the mean prediction and a 95\% confidence interval from Exact-GP using a zero-mean prior, revealing that the model is  underfitting due to the ground-truth non-zero mean, and the uncertainty quantification is severely impacted—\textbf{only 50\% of observed data is covered by the 95\% confidence interval.} In order to mitigate the underfitting issue, we suggest separating the mean prediction and uncertainty prediction. Especially, we propose a novel two-stage GPR model, illustrated in Figure \ref{fig:two stage GP}. In the first stage, a KRR is trained to capture the underlying data trends. In the second stage, the data is demeaned based on the mean predictions from the first stage, and a zero mean GPR is employed to model the uncertainty. This methodology is described in Algorithm \ref{alg:two-stage GP}. Separating mean and uncertainty predictions enables the zero mean GPR to avoid issues from a misspecified mean by using an accurate model to demean the data, aligning it with the zero mean GP assumption. As shown in Equation \eqref{eq:MEL}, MEL introduces a bias, causing zero mean GPR to underfit the data. In contrast, KRR does not underfit, as its estimator minimizing the mean square error loss is given by 
\(
\label{eq: prediction of KRR}
\hat{m}(\cdot) = \arg\min_{f \in \mathcal{H}_k} \left\{ \frac{1}{n} \sum_{i=1}^{n} (y_i - f(\mathbf{x}_i))^2 + \frac{\sigma_{\xi}^2}{n} \|f\|_{\mathcal{H}_k}^2 \right\},
\) where $\mathcal{H}_k$ is the reproducing kernel Hilbert space (RKHS) of kernel $k$.
In the rightmost panel of Figure \ref{fig:proposition-1-few-data-simplefunc}, the two-stage GP model is shown to capture the function more effectively, while also more accurately modeling the uncertainty. This observation further supports the adoption of the two-stage GPR approach over the Exact-GP. 

\begin{algorithm}
\caption{Two-stage GP model}
\begin{algorithmic}[1] 
\State \textbf{Input}: Observed training dataset $X_{\text{train}} = \{\bx_i, y_i\}_{i=1}^{N_{\text{train}}}$, testing dataset $X_{\text{test}} = \{\Tilde{\bx}_i\}_{i=1}^{N_{\text{test}}}$.
\State \textbf{Output}: Mean prediction $\hat{m}(\cdot)$ and variance prediction $\hat{\sigma}^2(\cdot)$ for $X_{\text{test}}$ .
\State Call Algorithm \ref{alg:automatic kernel searching} to find the optimal kernel $k_{\text{mean}}$ for mean prediction.
\State Train a KRR model with a fixed reguarlization parameter $\sigma_{\xi} = 0.01n$ on $X_{\text{train}}$ and get the mean prediction $\hat{m}(\cdot)$.
\State Compute the demeaned training labels $y_i^{\circ} = y_i - \hat{m}(\bx_i)$.
\State Call Algorithm \ref{alg:automatic kernel searching} to find the optimal kernel $k_{\text{var}}$ for $X_{\text{train}}^{\circ} = \{\bx_i, y_i^{\circ}\}_{i=1}^{N_{\text{train}}}$.
\State Train a zero-mean GP model $\mathcal{GP}(\bo, k_{\text{var}})$ on $X_{\text{train}}^{\circ}$ and get the variance prediction $\hat{\sigma}^2(\cdot)$.
\State \textbf{Return:} Mean prediction $\hat{m}(\cdot)$ and variance prediction $\hat{\sigma}^2(\cdot)$ for $X_{\text{test}}$.
\end{algorithmic}
\label{alg:two-stage GP}
\end{algorithm}


\section{Mitigating Kernel Misspecifications via Automatic Kernel Search}
\label{sec: kernel misspecification}
In the previous section, we introduced a two-stage GPR to address mean misspecification. Here, we assume a zero mean ground-truth function for convenience, as the two-stage GPR can handle mean misspecification. Another critical issue is kernel misspecification, which can compromise result reliability. While previous studies have shown that GPR can perform well despite benign kernel misspecification \cite{chowdhury2017kernelized, fiedler2021practical, Karvonen_2021, wang2022gaussian}, there remains a gap in systematically selecting the optimal kernel from a candidate set. We derive a probabilistic bound to assess kernel misspecification and propose an automatic kernel search method to select the kernel that best fits the data distribution, thereby enhancing model accuracy and applicability.
Our discussion begins with an analysis of prediction errors in KRR/GPR. For a new, unobserved data point \((\bx_*, y_*)\), where \(y_* = f(\bx_*) + \epsilon_{*}\), the prediction error is given by:
\(
|\hat{m}(\bx_*) - y_*| = |K_{*X}(K_{XX} + \sigma^2_{\xi}\mathbf{I})^{-1}\by_{n} - y_{*}|.
\)
This error can be decomposed into reducible and irreducible components:
\[
|\hat{m}(\bx_*) - y_*| = \underbrace{|K_{*X}(K_{XX} + \sigma^2_{\xi}\mathbf{I})^{-1}f_n - f(\bx_{*})|}_{\textbf{Reducible error}} + \underbrace{|K_{*X}(K_{XX} + \sigma^2_{\xi}\mathbf{I})^{-1}\mathbf{\epsilon_X} - \epsilon_{*}|}_{\textbf{Irreducible error}}
\]
where \(f_n\) is the vector of function values at the training points and \(\epsilon_X\) is the noise vector corresponding to these points.
In a well specified model with ideal hyperparameters, the \textbf{Reducible Error} approaches zero as the increase of sample size, as confirmed by Lemmas \ref{lemma: contarction-rates} and \ref{lemma: worst case error bound} in the Appendix \ref{appendix: supplementary Materials for sec3}. However, regardless of the kernel or parameters chosen, the \textbf{Irreducible error} is lower bounded probabilistically (Lemma \ref{lemma:tail lower bound for gaussian} in Appendix).
An intuitive explanation is that the \textbf{Irreducible error} corresponds to the uniform error in KRR for predicting a set of random noise \(\{\epsilon_i\}_{i=1}^{n}\), since
\(
K_{*X}(K_{XX} + \sigma^2_{\xi}\mathbf{I})^{-1}\mathbf{\epsilon_X} = \argmin_{f \in \mathcal{H}_{k}} \left(\frac{1}{n}\sum_{i=1}^{n}(\epsilon_i - f(\bx_i))^2 + \frac{\sigma_{\xi}^2}{n}\|f\|_{\mathcal{H}_{k}}\right).
\)
Actually, the following theorem shows the ratio of the prediction error to the irreducible error is upper bounded.
\begin{theorem}
\label{thm:model-misspecification-bound}
    Under mild assumptions (C1-C4 in Appendix~\ref{appendix: supplementary Materials for sec3}), suppose  $f\in \mathcal{H}_{k}$. Then with probability $1-\delta$ we have the following bound
    \begin{align}
        \frac{|\hat{m}_{n}(\bx_*) - y_*|}{|K_{*X}(K_{{XX}} + \sigma^2_{\xi}\mathbf{I})^{-1}\mathbf{\epsilon_X} -\epsilon_*|} \leq 1.1, \quad \forall x \in \Omega
   \label{eq:misbound}
    \end{align}
    when $n \geq \frac{\sigma_{\xi}^2}{\Big(\frac{0.1A(\delta)\sqrt{K_{*X}K_{X*}}}{(\lambda_{1} + \sigma_{\xi}^2)C'\|f\|_{\mathcal{H}_{k}}}\Big)^{\frac{2m_0}{2m_0 - d}}} $ where $A(\delta) = \sqrt{1 - 2\log{(1-\delta)}} - 1$, $\lambda_1$ is the largest eigenvalue of the kernel matrix $K_{XX}$ and $C'$ is the universal constant in Lemma \ref{lemma: contarction-rates}, $\hat{m}_{n}(\bx_*)$ is defined in Equation \eqref{eq: GPR-predictions} with a zero-mean prior.
\end{theorem}
Theorem \ref{thm:model-misspecification-bound} motivates us to propose a misspecification-checking algorithm based on the inequality \eqref{eq:misbound}. Although it's hard to estimate \(n\) in Theorem \ref{thm:model-misspecification-bound},  the ratio  $\frac{\textbf{Reducible error}}{\textbf{Irreducible error}}$ can quickly decay to a small quantity when the training size is not so small and the model is well specified or slightly misspecified.
Another unknown quantity in the bound is the ground-truth noise $\epsilon_i$ for each training data. In practice, this can be substituted with the predicted noise $\hat{\epsilon}_i = y_i - \hat{m}(\bx_i)$. 
With these simplifications, we propose an efficient algorithm for checking model misspecification and summarize it in Algorithm \ref{alg:check model misspecification}. 
\begin{algorithm}
\caption{Misspecification Checking Algorithm. }
\begin{algorithmic}[1] 
\State \textbf{Input}: Observed dataset $X$, kernel function $k$, desired threshold $p = 0.95$, a pre-defined desired probability $\delta$.
\State \textbf{Output}: Whether to reject hypothesis $H_0$: the model is well specified.
\For{$l=1$ \textbf{to} $100$}
    \State Call FPS sampling Algorithm \ref{alg:fps} to subsample a quasi-uniform dataset of size 500 from $X$, and split it into $X_{\text{train}}$ and $X_{\text{test}}$.
    \State Train the KRR/GP model on the training set until it converges.
    \State Compute the predicted means according to Equation \eqref{eq: GPR-predictions}.
    \State Compute $\Delta_i =  \frac{|\hat{m}_{n}(\bx_i) - y_i|}{|K_{*X}(K_{{XX}} + \sigma^2_{\xi}\mathbf{I})^{-1}\mathbf{\hat{\epsilon}_X}   - \hat{\epsilon}_*|} 
    $ for $\bx_i \in X_{\text{test}}$.
    \State Check if $\Delta_i \leq 1.1$ and compute empirical probability $\hat{P}_l = \frac{\#\{\Delta_i \leq 1.1\}}{|X_{\text{test}}|}$.
\EndFor
\State Compute the mean $\hat{P}$ from the 100 $\hat{P}_{l}$s. 
\State \textbf{Return}: Reject the hypothesis if $ \hat{P}< 1-\delta$; otherwise, accept the hypothesis.
\end{algorithmic}
\label{alg:check model misspecification}
\end{algorithm}

Finally, we introduce the Automatic Kernel Search Algorithm, designed to identify the optimal kernel for a specific task, as detailed in Algorithm \ref{alg:automatic kernel searching}.

\begin{algorithm}
\caption{Automatic Kernel Search Algorithm}
\begin{algorithmic}[1] 
\State \textbf{Input}: Observed dataset $X$, kernel functions dictionary 
$K = \{\text{kernel}:k(\cdot,\cdot)\}$.
\State \textbf{Output}: Optimal kernel $\text{kernel}^{*}$ from the dictionary.
\For{kernel, $k(\cdot,\cdot)$ \textbf{in} $K$}
    \State Initialize the best fitting probability as $p^* = 0$ and $\text{kernel}^{*} = \text{None}$.
    \State Run Algorithm \ref{alg:check model misspecification} for kernel function $k(\cdot, \cdot)$, which returns the empirical probability $p_{k}$.
    \If {$p_k > p^*$}
    \State Update $p^{*} = p_k$, $\text{kernel}^{*} = \text{kernel}$ 
    \EndIf
\EndFor
\State \textbf{Return:} $\text{kernel}^{*}$
\end{algorithmic}
\label{alg:automatic kernel searching}
\end{algorithm}
$\Delta_i$ in Line 7 of Algorithm \ref{alg:check model misspecification} is straightforward to compute for both KRR and GPR. A key distinction arises in Line 5 of Algorithm \ref{alg:check model misspecification}, where KRR typically employs cross-validation for training, whereas GPR utilizes the minimization of the NLL. The table below showcases the efficacy of Algorithm \ref{alg:automatic kernel searching} by comparing the performance of Exact-GP with and without the use of the Automatic Kernel Search (AKS) algorithm on various UCI regression datasets. Specifically, we partition each dataset into twenty folds same as \cite{han2022card}. We train two versions of Exact-GP: the baseline which consistently employs the RBF kernel, and the AKS-Exact-GP which selects a kernel from RBF, Mat\'ern-3/2, or Mat\'ern-1/2 by utilizing Algorithm \ref{alg:automatic kernel searching}. Performance metrics, including the mean and standard deviation of the NLL and Root Mean Square Error (RMSE), as defined in Equation \eqref{eq:scaled-likelihood} and  Section \ref{sec: numerical experiements for exact-GP} respectively, are computed across the twenty folds testing datasets for each dataset. For the majority of datasets evaluated, AKS-Exact-GP consistently outperforms the standard Exact-GP utilizing the RBF kernel. Notably, in the case of the Power dataset, the NLL for the Exact-GP with the RBF kernel is significantly higher compared to that of AKS-Exact-GP, despite the RMSE values appearing normal. This suggests that the uncertainty quantification performed by the Exact-GP with the RBF kernel is likely probelmatic. \textbf{Additionally, the high NLL values observed for the Naval dataset with both methods underscore the issue discussed in Section \ref{sec: mean specification}. This not only highlights the significant impact of mean misspecification on UQ but also advocates for the adoption of a two-stage GP framework, since well specified mean and kernel are equally important for the prediction task requiring UQ.}
\begin{table}[h]
\centering
\caption{RMSE and NLL metrics for various datasets using Exact-GP with RBF kenrel and AKS-Exact-GP.}
\label{tab:kernel checking metrics}
\renewcommand{\arraystretch}{1.0} 
\setlength{\tabcolsep}{3.0pt} 
\begin{tabular}{@{}lcccccc@{}}
\toprule
\textbf{Dataset} & \multicolumn{2}{c}{Data Info (/Fold)} & \multicolumn{2}{c}{\textbf{RBF}} & \multicolumn{2}{c}{\textbf{AKS}} \\
\cmidrule(lr){2-3} \cmidrule(lr){4-5} \cmidrule(lr){6-7}
 & $n$ & $d$ & RMSE & NLL & RMSE & NLL \\ \midrule
 Yacht    & $277$ & $6$ & $\mathbf{2.12\pm 1.23}$ & $-1.03\pm0.15$  &$2.94\pm 1.23 $ & $\mathbf{-0.91\pm0.12}$ \\
Boston    & $455$ & $13$ & $\mathbf{2.82\pm 0.70}$ & $ -0.44\pm 0.22$  &$2.87\pm 0.71$ & $\mathbf{-0.66\pm0.03}$ \\
Energy   & $691$ & $8$ & $\mathbf{1.53\pm 0.66}$ & $\mathbf{-0.84\pm0.10}$  &$2.16\pm 0.25$& $-0.74\pm0.01$ \\
Concrete  & $927$ & $8$ & $ 5.30\pm 0.58$ & $ -0.49\pm0.13$ & $\mathbf{5.26\pm 0.51}$&$ \mathbf{-0.58\pm0.09}$ \\
Wine    & $1439$ & $11$ &  $0.60\pm 0.05$ & $0.17\pm 0.83$ &  $\mathbf{0.57\pm 0.05}$&$\mathbf{ -0.60\pm0.40}$ \\
Kin8nm     & $7373$ & $8$ & $ \mathbf{0.07\pm 0.00}$&$ -1.09\pm0.58$  & $\mathbf{0.07\pm 0.00}$& $\mathbf{-1.15\pm0.78}$ \\
Power     & $8611$ & $4$ & $3.64\pm 0.22$& $\textcolor{red}{1555\pm11990}$ & $\mathbf{3.52\pm 0.18}$& $\mathbf{-0.49\pm0.79}$ \\
Naval     & $10741$ & $16$ & $\mathbf{0.00\pm 0.00}$& $\textcolor{red}{2017\pm 6202}$ &$\mathbf{0.00\pm 0.00}$& $\mathbf{1326\pm5713}$ \\
\bottomrule
\end{tabular}
\end{table}

\section{Hyperparameters Tuning via Subampling Warm Start}
\label{sec: subsampling GP}
Having addressed mean and kernel misspecification, we now focus on the final potential source of error: inexact training of GPR. Evaluating the NLL or its gradient incurs a computational cost of $O(n^3)$, making the training of GPR expensive for large-scale datasets. A common approach is to train GPR on a subsampled dataset and then apply the learned hyperparameters to the full dataset for prediction. We will demonstrate empirically in Figure \ref{fig:UCI wine contour plot} that this approach typically results in a suboptimal solution near a local minimum. Conversely, we argue both theoretically and empirically in the following sections that hyperparameters optimized on subsampled data provide an effective initialization for full-scale GPR training.
Before we proceed to the main theorem, we need the following assumption:
\begin{itemize}
    \item \textbf{C5}: Suppose the mean and the functional form of the kernel of GPR is well specified, and there exists an oracle algorithm $\mathcal{O}$ which could inform us of the optimal hyperparameters $\theta^*$ of $EL(\theta)$. 
\end{itemize}
\begin{remark}
Though this assumption may initially appear far-fetched, it effectively abstracts the typical application of GPR in practice. We discussed this assumption in details in Appendix \ref{sec: subsampling GP}.
\end{remark}
\begin{theorem}
\label{thm: subsampled GP}
For each of the lengthscale $l$, outputscale $\sigma_f$ or noise parameter $\sigma_{\xi}$ with the other two parameters fixed, the optimal hyperparameters $\theta^*$ obtained using the oracle algorithm $\mathcal{O}$ either from a subsampled dataset or a full dataset are the same, and it is a local minimum for $EL(\theta;n)$ independent of the size of training data $n$ if the training data follows the GP assumptions. Furthermore, suppose $\theta^*$ is minimal point in the domain $\Omega$ and $L({\theta;n})$ is used in training instead of the $EL(\theta;n)$, then $\theta^*$ is a $\epsilon$-minimum, i.e., $L(\theta^{*};n) \leq L(\theta;n) + \epsilon(K_{n}(\theta),K_n(\theta^*, \delta)), ~\forall \theta \in \Omega$ with probability $1-\delta$ for any $\delta \in (0, 1)$, where $\epsilon = 2\max{\Big\{ \frac{\sqrt{2C\|\Tilde{K}\|_{F}\log{\frac{2}{\delta}}}}{n}, \frac{2C\|\Tilde{K}\|\log{\frac{2}{\delta}}}{n}
 \Big\}}$, where $\Tilde{K}(\theta) = U_{*}^{\top}K^{-1}U_{*}$ and  $ K(\theta^{*}) = Q_{*} \Lambda_{*} Q_{*}^{\top} = (Q_{*}\Lambda_{*}^{\frac{1}{2}})(Q_{*}\Lambda_{*}^{\frac{1}{2}})^{\top}:=U_{*}U_{*}^{\top}$, $Q_{*} \Lambda_{*} Q_{*}^{\top}$ is the eigenvalue decomposition of $K_{*}$ with orthonormal eigenvectors $U_{*}$.
\end{theorem}
In Figure \ref{fig:UCI wine contour plot}, we provide empirical support for Theorem \ref{thm: subsampled GP} using the UCI wine dataset. Initially, approximately 10\% of the dataset is used to determine the optimal hyperparameters—lengthscale, output scale, and noise—by minimizing the NLL with initial parameters \(\theta_0 = (0.693, 0.693, 0.693)\) as per GPyTorch defaults. We then visualize the NLL contours around these optimal hyperparameters on the entire dataset. This process is repeated with larger subsets (50\%, 90\%, and 100\% of the data) using the same \(\theta_0\). The contour plots show that the optimal hyperparameters from the training sets closely align with those for the entire dataset. As the training data size increases, the derived optimal hyperparameters better approximate the optimum, validating the theorem. Further experiment details and validations are in Appendix \ref{appendix: supplementary Materials for sec4}. As suggested by Theorem \ref{thm: subsampled GP} and this example, subsampling with insufficient data can lead to hyperparameter misspecification. \textbf{Thus, we recommend using the optimal hyperparameters from a small randomly subsampled subset as a warm start for further large-scale training on full dataset. This strategy can efficiently reduce the cost via training exclusively on full dataset while avoiding hyperparameter misspecification caused by training only on subsampled datasets.}

\newsavebox{\diamondder}
\savebox{\diamondder}{%
\begin{tikzpicture}
    \fill[red] (0,0) -- (1.5mm,1.5mm) -- (3mm,0) -- (1.5mm,-1.5mm) -- cycle;
\end{tikzpicture}%
}
\newsavebox{\cross}
\savebox{\cross}{%
\begin{tikzpicture}
 \begin{scope}[shift={(2,0)}]
        \draw[red, thick] (-1.5mm,-1.5mm) -- (1.5mm,1.5mm); %
        \draw[red, thick] (-1.5mm,1.5mm) -- (1.5mm,-1.5mm); %
    \end{scope}
    \end{tikzpicture}%
}  
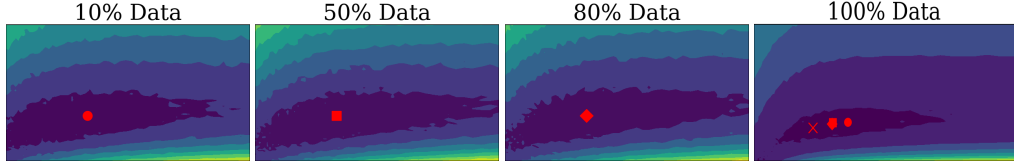
\begin{figure}[!htp]
\centering
\includegraphics[width=.23\textwidth]{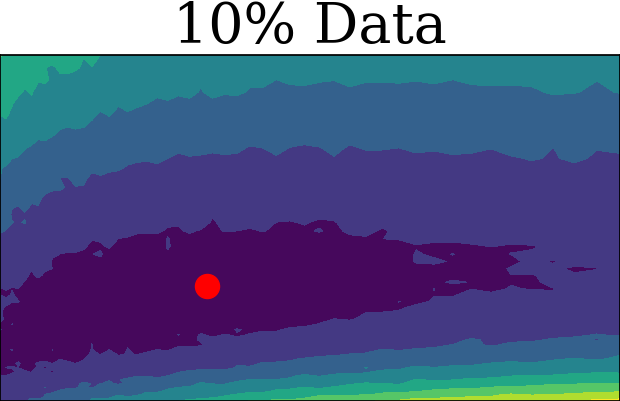}
\includegraphics[width=.23\textwidth]{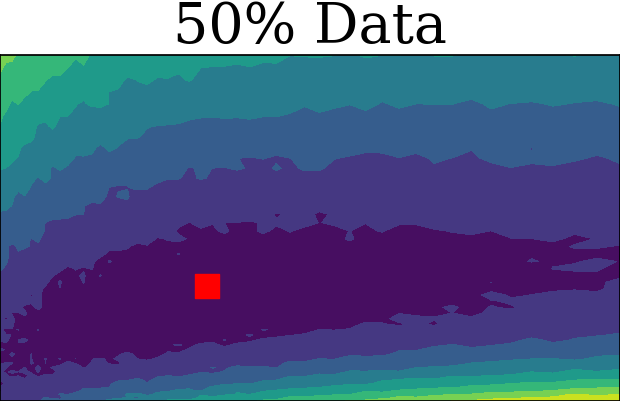}
\includegraphics[width=.23\textwidth]
{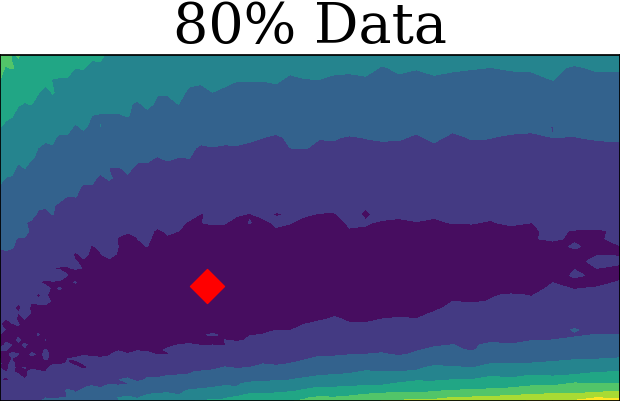}
\includegraphics[width=.245\textwidth, height=2.10cm]
{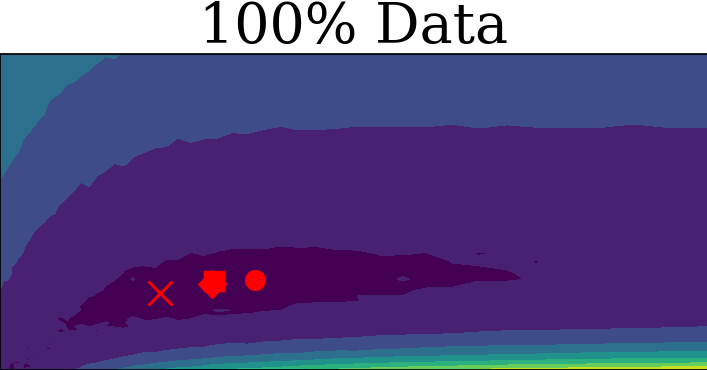} 
    \caption{\textbf{Contour Plot of NLL for the UCI Wine Dataset}: This plot illustrates the pairwise contours plot around the optimal lengthscale vs. noise. Optimal hyperparameters trained on 10\%, 50\%, 80\%, and 100\% datasets are marked with red dot
    \begin{tikzpicture}
    \fill[red] (0,0) circle (3pt); 
    \end{tikzpicture}
    , square
    \begin{tikzpicture}
    \fill[red] (0,0) rectangle +(2mm,2mm);
    \end{tikzpicture}
    , diamond
    \protect\usebox{\diamondder}
    , and cross
    \protect\usebox{\cross}
    , respectively. Lighter areas indicate higher NLL values, while darker areas signify lower NLL values. All the contour plots are plotted on full training datasets.}
\label{fig:UCI wine contour plot}
\end{figure}

\section{Two Approaches for GP}
\label{sec: Two approaches for GP}
Based on the the proposed two-stage GP framework as in Figure \ref{fig:two stage GP}, we propose two distinct approaches tailored to different application needs. For scenarios where computational resources are constrained and the primary interest lies in mean prediction, we recommend utilizing scalable GP methods such as GPNN \cite{Allison_Stephenson_F_Pyzer-Knapp_2023}, NNGP \cite{Datta_2022}, and SVGP \cite{hensman2013gaussian} within our two-stage GPR framework. In this paper, we select GPNN \cite{Allison_Stephenson_F_Pyzer-Knapp_2023} as our baseline due to its simplicity, robustness, and effectiveness. The implementation details are presented in Algorithm \ref{alg: Combined Scalable GP} Conversely, when computational power is abundant and UQ is crucial—such as in safety-critical applications including robotic control \cite{fiedler2021practical} and medical diagnostics—Exact-GP \cite{lu2024uncertainty} is the preferred choice since Exact-GP utilizes the full training dataset, resulting in higher accuracy. The complete two-stage Exact-GP algorithm is detailed in Algorithm \ref{alg: Combined Exact GP}.


\section{Numerical results}
In this section, we conduct three sets of experiments to evaluate our framework. First, we assess the performance on various small-scale UCI regression datasets, comparing Exact-GP, two-stage Exact-GP based on several popular metrics. Next,  we introduce a novel uncertainty-aware metric to measure the performance of UQ of an estimator for regression tasks. Then we demonstrate the UQ performance of the two-stage GPNN for several large-scale datasets based on this new metric. Finally, we compare the UQ capabilities of two-stage Exact-GP and Exact-GP on a safe-critical application, demonstrating the superiority of the two-stage Exact-GP approach on classification tasks. For slef-completeness, we include a brief introduction on Gaussian Process Classification (GPC) in Appendix \ref{appendix: GPC}.
\label{sec: numerical experiments}
\subsection{Performance Comparison for Exact-GP on UCI Dataset}
\label{sec: numerical experiements for exact-GP}
In this section, we compare the performance of Exact-GP and two-stage Exact-GP on small-scale UCI datasets. For these datasets, we report the exact NLL defined in Equation \eqref{eq:scaled-likelihood}, an effective metric for UQ. Additionally, we report \textbf{RMSE (Root Mean Square Error)}, defined as \(\sqrt{\frac{1}{n} \sum_{i=1}^{n} (\hat{y}_i - y_i)^2}\), and \textbf{QICE (Quantile Interval Coverage Error)}, calculated as \(\frac{1}{M} \sum_{m=1}^M \left| r_m - \frac{1}{M} \right|\). Here, \(r_m := \frac{1}{N} \sum_{n=1}^N \mathbf{1}_{y_n \geq \hat{y}_n^{m-1}} \cdot \mathbf{1}_{y_n \leq \hat{y}_n^{m}}\) represents the proportion of true targets falling between the \(m-1\)-th and \(m\)-th quantiles of predicted targets. This metric, cited from \cite{han2022card}, evaluates prediction calibration, serving as an important measure of the model's UQ capability.

\textit{For all three metrics, a smaller value indicates better model performance.} We report the performance of Exact-GP with a default RBF kernel and Two-stage Exact-GP defined in Section \ref{sec: Two approaches for GP} in the follwing Table \ref{tab:Exact-GP metrics}.
\begin{table}[htbp]
\centering
\caption{RMSE and NLL metrics for various datasets using Exact-GP with RBF kenrel and AKS-Exact-GP.}
\label{tab:Exact-GP metrics}
\renewcommand{\arraystretch}{1.0} 
\setlength{\tabcolsep}{2.0pt} 
\begin{tabular}{@{}lcccccccc@{}}
\toprule
\textbf{Dataset} & \multicolumn{3}{c}{\textbf{Exact-GP(RBF)}} & \multicolumn{3}{c}{\textbf{Two-stage Exact-GP}} \\
 \cmidrule(lr){2-4} \cmidrule(lr){5-7}
 & RMSE & NLL &QICE & RMSE & NLL &QICE \\ \midrule
Yacht   & $1.29\pm 0.42$ & $-1.15\pm0.03$ & $6.37\pm 1.47$ & $\mathbf{0.41\pm 0.16}$ & $\mathbf{-1.52\pm0.04}$ & $\mathbf{5.03\pm 1.59}$\\
Boston    & $2.77\pm 0.68$ & $\mathbf{-0.22\pm0.05}$ & $3.78\pm 1.00$  & $\mathbf{2.70\pm 0.67}$ & $1.11\pm1.49$ & $\mathbf{3.55\pm 0.93}$\\
Energy   & $0.89\pm 0.12$ & $-0.93\pm0.03$ & $3.36\pm 0.94$  & $\mathbf{0.37\pm 0.07}$ & $\mathbf{-1.51\pm0.03}$ & $\mathbf{2.24\pm 0.56}$ &\\
Concrete & $5.35\pm 0.64$ & $\mathbf{-0.39\pm0.07}$ & $2.88\pm 0.67$ & $\mathbf{3.78\pm 0.58}$ & $0.03\pm0.50$ & $\mathbf{2.23\pm 0.57}$   \\
Wine   & $0.62\pm 0.04$ & $0.95\pm0.05$ & $\mathbf{13.21\pm 0.31}$ & $\mathbf{0.60\pm 0.04}$ & $\mathbf{0.67\pm0.78}$ & $\mathbf{13.2\pm 0.31}$ \\
Kin8nm  &$\mathbf{0.07\pm 0.00}$ & $\mathbf{-1.03\pm0.21}$ & $0.95\pm 0.24$  &$\mathbf{0.07\pm 0.00}$ & $-0.15\pm0.03$ & $\mathbf{0.94\pm 0.27}$  \\
Power  & $3.75\pm 0.19$ & $\textcolor{red}{3111\pm16814}$ & $\mathbf{1.05\pm 0.27}$& $\mathbf{3.23\pm 0.20}$ & $\mathbf{0.06\pm0.26}$ & $15.63\pm 0.09$\\
Naval  & $\mathbf{0.00\pm 0.00}$ & $\textcolor{red}{924.1\pm3892}$ & $0.97\pm 0.43$& $\mathbf{0.00\pm 0.00}$ & $\mathbf{-1.62\pm0.00}$ & $\mathbf{0.69\pm 0.22}$ & \\
\bottomrule
\end{tabular}
\end{table}
Two-stage GP outperforms Exact-GP on most of the datasets . It's worth noting that for Power and Naval datasets, Exact-GP reports very large NLL compared to Two-stage Exact-GP, which usually indicates a unreliable UQ.

\subsection{Uncertainty Quantification for Scalable GP on UCI datasets}
\label{sec: UQ for GPNN}
In this section, we introduce the Uncertainty-Aware RMSE (UA-RMSE), a metric designed to evaluate the quality of UQ provided by estimators. In Gaussian Process (GP) models, UQ is typically represented as a standard error estimate for predictions. For each prediction \(\hat{y}_{i} = \hat{m}(\mathbf{x}_i)\), the associated standard error is \(\hat{s}_i = \sqrt{\hat{k}(\mathbf{x}_i, \mathbf{x}_i)}\) as described in Equation \eqref{eq: GPR-predictions}. We define two variations of UA-RMSE based on the certainty of predictions: High Certainty RMSE (HC-RMSE) and Low Certainty RMSE (LC-RMSE). HC-RMSE is calculated as the RMSE for predictions where the standard error is lower than the \(100q\)-th quantile of the whole standard error estimates on test data denoted as \(\hat{s}_{q}\), whereas LC-RMSE is calculated for predictions where the standard error exceeds the \((1-q)100\)-th quantile of that denoted as \(\hat{s}_{1-q}\) where $q\in (0,1)$. These metrics provide insights into the model's performance across varying levels of uncertainty.
We formally define HC-RMSE and LC-RMSE as follows:
\(
\text{HC-RMSE} := \sqrt{\frac{1}{qn} \sum_{h=1}^{qn} (\hat{y}_h - y_h)^2}, \quad \text{where } \hat{s}_{h} \leq \hat{s}_{q},
\)
\(
\text{LC-RMSE} := \sqrt{\frac{1}{qn} \sum_{l=1}^{qn} (\hat{y}_l - y_l)^2}, \quad \text{where } \hat{s}_{l} > \hat{s}_{1-q}.
\)
We split the test instances into ``[Method]-certain'' and ``[Method]-uncertain'' groups. A model with high-quality UQ is expected to exhibit a lower HC-RMSE compared to LC-RMSE. The lower the HC-RMSE, the more effective the model is at quantifying uncertainties. In the following bar plot, we plot the HC-RMSE and LC-RMSE for both GPNN and two-stage GPNN on seven UCI datasets with $q=0.1$.
Two-stage GPNN achieves lower RMSE for high certainty instances over all seven datasets. For the discrepancy between the HC-RMSE and LC-RMSE, two-stage GPNN also beats GPNN over 5 datasets. The details of the experiments and information of datasets can be found in Appendix \ref{appendix: supplementary Materials for sec6}.
\begin{figure}[htbp!]
 \centering
 \begin{tikzpicture}
\begin{axis}[
    width=0.95\textwidth, 
    height=4cm, 
    ybar=4.5pt,
    ymin=0, ymax=1,
    enlarge x limits=0.10, 
    ylabel={RMSE},
    ymajorgrids=true,
    symbolic x coords={Poletele, Bike, Protein, Ctslice, Road3d, Song, HouseE},
    xtick=data,
    xticklabel style={rotate=45, anchor=east, font=\small},
    nodes near coords,
    nodes near coords style={
        font=\tiny,
        /pgf/number format/.cd,
        fixed,             
        sci subscript,     
        precision=2,
        fixed zerofill     
    },
    bar width=4.5pt,
    legend columns=4, 
    legend style={
        font=\normalsize, 
        fill=none,
        at={(0.5,1.03)}, 
        anchor=south, 
        draw=none 
    },
       ticklabel style={font=\large},
    every axis plot/.append style={fill},
    cycle list/Paired,
]
\addplot[index of colormap=1 of Paired] coordinates {(Poletele, 0.14) (Bike, 0.81) (Protein, 0.47) (Ctslice, 0.01) (Road3d, 0.10) (Song, 0.61) (HouseE, 0.03)};
\addplot[index of colormap=0 of Paired] coordinates {(Poletele, 0.28) (Bike, 0.51) (Protein, 0.71) (Ctslice, 0.33) (Road3d, 0.30) (Song, 0.79) (HouseE, 0.07)};
\addplot[index of colormap=3 of Paired] coordinates {(Poletele, 0.14) (Bike, 0.39) (Protein, 0.19) (Ctslice, 0.01) (Road3d, 0.07) (Song, 0.58) (HouseE, 0.01)};
\addplot[index of colormap=2 of Paired] coordinates {(Poletele, 0.26) (Bike, 0.51) (Protein, 0.77) (Ctslice, 0.37) (Road3d, 0.13) (Song, 0.80) (HouseE, 0.10)};
\legend{GPNN-certain, GPNN-uncertain, 2StGPNN-certain, 2StGPNN-uncertain}
\end{axis}
\end{tikzpicture}
 \vspace{-0.6cm}
 \caption{Uncertainty quantification results in RMSE.}
 \label{fig:uq_UCI}
\end{figure}
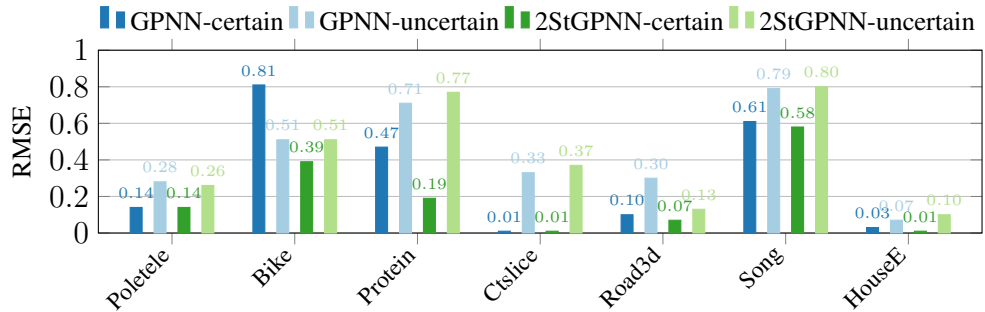
It's also worth mentioning that the UQ of GPNN for the Bike dataset is problematic, since the GPNN-certiain RMSE is much larger than the GPNN-unceratin. This kind of problematic UQ may lead to diastrous result for safe-critical applications like the ones in next section.

\subsection{Uncertainty Quantification for GP-Enhanced Pre-Trained Foundation Models}
\label{sec: UQ for PFM}
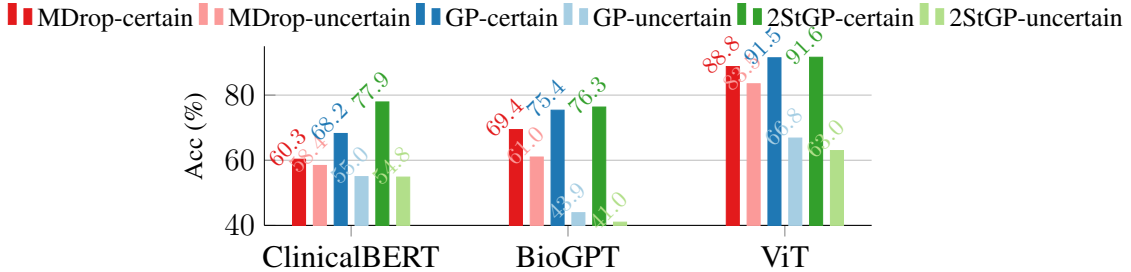
\begin{figure}[htbp!]
 \centering
 \begin{tikzpicture}[scale=0.98]
\begin{axis}[
    width=0.7\textwidth, 
    height=4cm, 
    ybar=3pt,
    ymin=40, ymax=95,
    axis lines*=left,
    enlarge x limits=0.2,
    ylabel={Acc (\%)},
    ymajorgrids=true,
    symbolic x coords={ClinicalBERT, BioGPT, ViT},
    xtick=data,
    nodes near coords,
    nodes near coords style={rotate=45,font=\footnotesize,/pgf/number format/.cd,fixed zerofill,precision=1},
    bar width=5pt,
    legend columns=6, 
    legend style={
        font=\normalsize, 
        fill=none,
        at={(0.5,1.03)}, 
        anchor=south, 
        draw=none 
    },
    ticklabel style={font=\large},
    every axis plot/.append style={fill},
    cycle list/Paired,
]
\addplot[index of colormap=5 of Paired] coordinates {(ClinicalBERT, 60.32) (BioGPT, 69.41) (ViT, 88.75)};
\addplot[index of colormap=4 of Paired] coordinates {(ClinicalBERT, 58.40) (BioGPT, 60.99) (ViT, 83.48)};
\addplot[index of colormap=1 of Paired] coordinates {(ClinicalBERT, 68.22) (BioGPT, 75.37) (ViT, 91.46)};
\addplot[index of colormap=0 of Paired] coordinates {(ClinicalBERT, 54.97) (BioGPT, 43.90) (ViT, 66.81)};
\addplot[index of colormap=3 of Paired] coordinates {(ClinicalBERT, 77.90) (BioGPT, 76.33) (ViT, 91.62)};
\addplot[index of colormap=2 of Paired] coordinates {(ClinicalBERT, 54.83) (BioGPT, 40.99) (ViT, 62.97)};
\legend{MDrop-certain, MDrop-uncertain, GP-certain, GP-uncertain, 2StGP-certain, 2StGP-uncertain}
\end{axis}
\end{tikzpicture}
 \vspace{-0.6cm}
 \caption{Uncertainty quantification results in accuracy (\%).}
 \label{fig:uq_PFM}
\end{figure}

Pre-trained foundation models (PFMs) have demonstrated impressive performance in a wide range of prediction tasks. PFMs are mostly large Transformer-based models~\cite{radford2019gpt2} with minimum modification and are pre-trained in a self-supervised manner with large unlabeled data. When adapting PFMs to downstream applications, the common practice is to freeze the learned parameters of all neural network layers except for \textbf{the last layer}. The default implementation of the last layer is the fully connected layer (linear transformation with optional non-linear activation), resulting in a notable limitation that lies in lacking the ability to express predictive uncertainty~\cite{lu2024uncertainty}. 
We propose to utilize our two-stage GP model to serve as the last prediction layer for PFMs, thus providing valuable uncertainty quantification for safety-critical applications. 
Figure~\ref{fig:uq_PFM} shows the uncertainty quantification experimental results, when using the proposed two-stage GP as a plug-in module for various types of PFMs. We select three PFMs as the backbone models: ClinicalBERT~\cite{wang2023clinicalbert}, BioGPT~\cite{luo2022biogpt}, and ViT~\cite{dosovitskiy2020vit}. For UQ methods, we compare the Monte Carlo Dropout~\cite{gal2016mcdropout}, Exact-GP, with our proposed two-stage Exact-GP. It is worth noting that we utilize the Dirichlet classification likelihood trick~\cite{milios2018dirichlet} to transform the classification targets into the regression targets that fit the requirement of GP regression method.
Following the setting introduced in \cite{han2022card,lu2024uncertainty}, we conduct paired t-test to split the predicted categories of test instances into ``[Method]-certain'' and ``[Method]-uncertain'' groups. We expect high accuracy in certain group, and low accuracy in uncertain group. That is because uncertain predictions would be sent to clinicians for further verification in real clinical practice, and we intend to send mainly wrong predictions for double verification to relief clinician's burden.
For ClinicalBERT and BioGPT, we test them on the MedNLI~\cite{romanov2018mednli} dataset targeting on identifying potential clinical outcomes based on past medical history stored in the textual clinical notes. For Vit, we test them on the BreakHis~\cite{spanhol2015breakhis} dataset targeting on distinguishing benign tumors and malignant tumors in breast cancer histopathological images. Both datasets are safety-critical patient risk prediction tasks where uncertainty quantification is indispensable. As can be seen from Fig.~\ref{fig:uq_PFM}, our proposed two-stage Exact-GP consistently outperforms the other two baseline methods, by obtaining higher accuracy on certain predictions and lower accuracy on uncertain predictions.

\section{Limitations}
\label{sec:limitations}
The KRR in the first stage of the Two-Stage GPR framework (Section \ref{sec: mean specification}) can be replaced with other methods, though we haven't empirically studied this possibility. 
We also include a GP classification example in Section \ref{sec: UQ for PFM} using the approach in \cite{milios2018dirichlet}, detailed in Appendix \ref{appendix: background}. Although the Two-Stage GPC outperforms GPC, it shows only marginal improvement on the BioGPT and ViT models. This may be due to the approximation in the GPC via the GPR approach (Appendix \ref{appendix: background}). Future work could develop a two-stage GP model specifically for classification tasks.

\section{Conclusion}
\label{sec:conclusion}
In this study, we addressed several misspecifications encountered in applying Gaussian Processes (GP) to real-world datasets. We introduced a two-stage GPR framework to mitigate mean misspecification and enhance uncertainty quantification (UQ). Additionally, we developed an automatic kernel search algorithm to correct kernel misspecifications and proposed a subsampling warm start strategy to efficiently address hyperparameter misspecification.
These advancements led to the development of two tailored approaches: the two-stage scalable GP and the two-stage Exact-GP. Our findings indicate that the two-stage GP framework significantly improves UQ over traditional GP models. For scenarios requiring precise UQ and ample computational resources, we recommend the two-stage Exact-GP. Conversely, for contexts where mean prediction is crucial or resources are limited, the scalable two-stage GPNN provides a viable solution, offering better UQ than baseline models without compromising prediction accuracy.

\bibliographystyle{plain}
\bibliography{reference}

\clearpage
\appendix
\section{Background Materials}
\label{appendix: background}
\subsection{Popular kernels}
\label{subsec:kernels}
Here are several kernels we used in this paper.

\begin{itemize}
    \item \textbf{RBF Kernel}: 
                            $k(\bx_i,\bx_j) = \sigma_{f}^2 \exp\left(-\frac{\|\bx_i - \bx_j\|^2}{l^2}\right)$
    \item \textbf{Mat\'ern-3/2 Kernel}:
                            $ k(\mathbf{x}_i, \mathbf{x}_j) = \sigma_{f}^2 \left(1 + \frac{\sqrt{3} \|\mathbf{x}_i - \mathbf{x}_j\|}{l}\right) \exp\left(-\frac{\sqrt{3} \|\mathbf{x}_i - \mathbf{x}_j\|}{l}\right)$
    \item \textbf{Mat\'ern-1/2 Kernel}:
                            $k(\mathbf{x}_i, \mathbf{x}_j) = \sigma_{f}^2 \exp\left(-\frac{\|\mathbf{x}_i - \mathbf{x}_j\|}{l}\right)$  
\end{itemize}
\subsection{Gaussian Process Classification (GPC)}
\label{appendix: GPC}
We include an example of a safety-critical application in Section \ref{sec: UQ for PFM}. Since our framework is based on GPR, we adopted the approach from \cite{milios2018dirichlet} to treat GPC as GPR via Dirichlet likelihood. For completeness, we briefly introduce this method.

In a multi-classification problem with \(k\) classes, we use a Dirichlet distribution as the likelihood model \(p(y|\pi) = \text{Cat}(\pi)\), where \(\pi \sim \text{Dir}(\mathbf{\alpha})\) and \(\text{Dir}(\mathbf{\alpha})\) is a Dirichlet distribution with \(\mathbf{\alpha} = (\alpha_1, \dots, \alpha_k)\). It is well known that \(x \sim \text{Dir}(\alpha)\) can be generated using \(k\) Gamma distributions \(\Gamma(\alpha_i, 1)\), which are approximated via \(\tilde{x} \sim \text{LogNormal}(\tilde{y}_i, \tilde{\sigma}_i^2)\), where \(\tilde{y}_i = \log{\alpha_i} - \frac{\tilde{\sigma}_i^2}{2}\) and \(\tilde{\sigma}_i = \log{\left(\frac{1}{\alpha_i} + 1\right)}\).

Thus, we can use a Gaussian likelihood \(p(\tilde{y}_i|\mathbf{f}) = N(\mathbf{f}, \tilde{\sigma}_f^2)\) in the log-space, regarding \(\tilde{y}_i\) as the transformed continuous targets for GPR. Here, \(\mathbf{f} = [f_1, \dots, f_k]\) are \(k\) latent GPs. By using the lognormal approximation of the Gamma distribution, we replace the intractable Dirichlet likelihood with a tractable Gaussian likelihood. For prediction, we marginalize over the latent process. Specifically, since \(\tilde{x}\) is in log-space, we exponentiate it to treat it as a Gamma distribution and use these Gamma distributions to construct samples for Dirichlet distributions. The probability for class \(j\) can be computed as follows:

\begin{align*}
    p[j|X, \mathbf{x}_{*}] = \int \frac{e^{f_j^*}}{\sum_{i=1}^{k} e^{f_i^*}} p(\mathbf{f}|X, \mathbf{x}_*) d\mathbf{f}
\end{align*}

where \(p(\mathbf{f}|X, \mathbf{x}_*)\) is the posterior distribution of the latent GP \(f_j\). This can be further approximated by sampling \(N\) times from \(p(\mathbf{f}|X, \mathbf{x}_*)\):

\begin{align*}
    p[j|X, \mathbf{x}_{*}] \approx \frac{1}{N} \sum_{l=1}^{N} \frac{e^{f_j^{*,l}}}{\sum_{i=1}^{k} e^{f_i^{*,l}}}
\end{align*}

\section{Supplementary Materials for Section \ref{sec: mean specification}}
\label{appendix: supplementary Materials for sec2}
Authors in \cite{malinin2018predictive} separated the distributional uncertainty from epistemic uncertainty and decomposed the total uncertainty of a model into data uncertainty, distributional uncertainty, and model uncertainty. Data uncertainty or aleatoric uncertainty is referred to as the known-unknown, which is handled by the likelihood model in Gaussian Process Regression (GPR), while the distributional uncertainty is effectively managed by the kernel due to its distance-awareness \cite{liu2020simple}, i.e., when the testing data has a different distribution from the training set thus the testing points are far away from the training points. Finally, the model uncertainty is taken care of by the specification of the kernel function. Therefore, with the wrong mean specification, the uncertainty quantification will be impacted due to the obtained wrong hyperparameters. This suggests separating the mean prediction and UQ is critical for UQ.\\


 \begin{proof}[Proof of Theorem \ref{thm: misspecification of mean}]
    The gradient of the \(MEL\) can be computed as follows,
    \begin{align}
    \mathbb{E}[\frac{\partial L(\theta;X_n)}{\partial \theta}] &=  \frac{1}{2n}\mathbb{E}[\Tr{K_{n}^{-1}(\bI_{n} - (\by-m(X))(\by_{n}-m(X))^{\top}K_{n}^{-1})\frac{\partial K_n}{\partial \theta}}] \\
    & + \Tr{K_{n}^{-1}m(X)m(X)^{\top}K_{n}^{-1}\frac{\partial K_n}{\partial \theta}}\\
    & = \frac{1}{n}\Tr{K_{n}^{-1}(\bI_{n} - K_{n}(\theta^{*}) K_{n}^{-1})\frac{\partial K_n}{\partial \theta}}\\
    & + \Tr{m(X)^{\top}K_{n}^{-1}\frac{\partial K_n}{\partial \theta} K_{n}^{-1}m(X)}
\end{align}
\end{proof}
\textbf{Experimental Details of Figure 2}:
Exact-GP and two-stage GP both use a AdamW optimizer and $lr=0.1$. We also found the lr scheduler get\_cosine\_with\_hard\_restarts\_schedule\_with\_warmup from transformer package can make the training of GP more stable thus is used throughout all our experiments. This figure can be reproduced with propositionkrr.py in our github repo. \\
Here we also provide another a supplementary example to Figure \ref{fig:proposition-1-few-data-simplefunc}.
\begin{figure}[!htp]
    \centering
    \includegraphics[width=.28\textwidth]{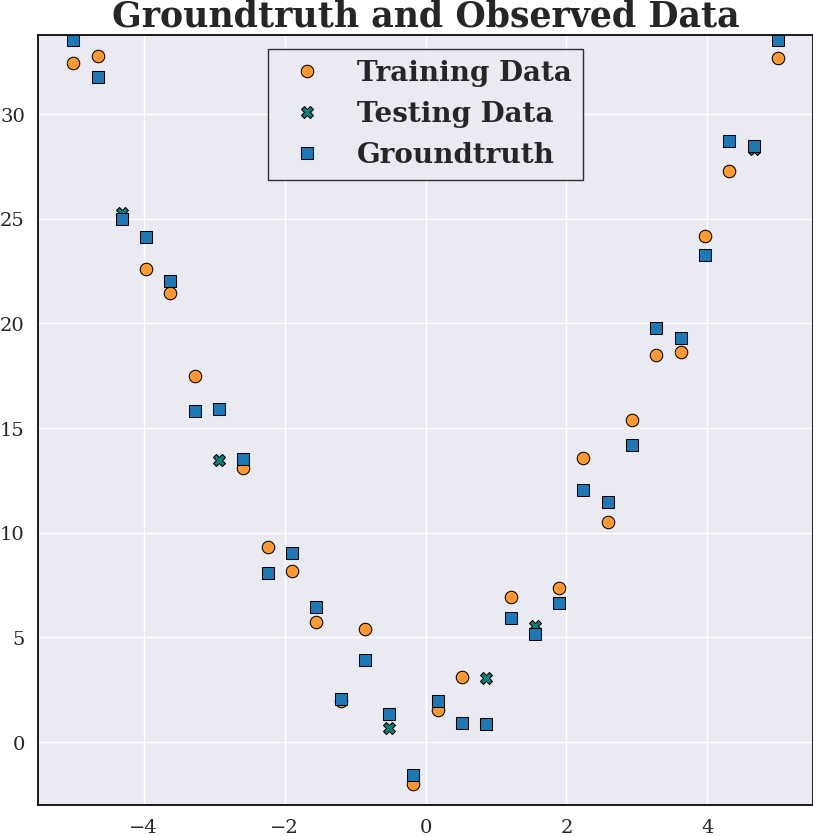}
    \includegraphics[width=.28\textwidth]{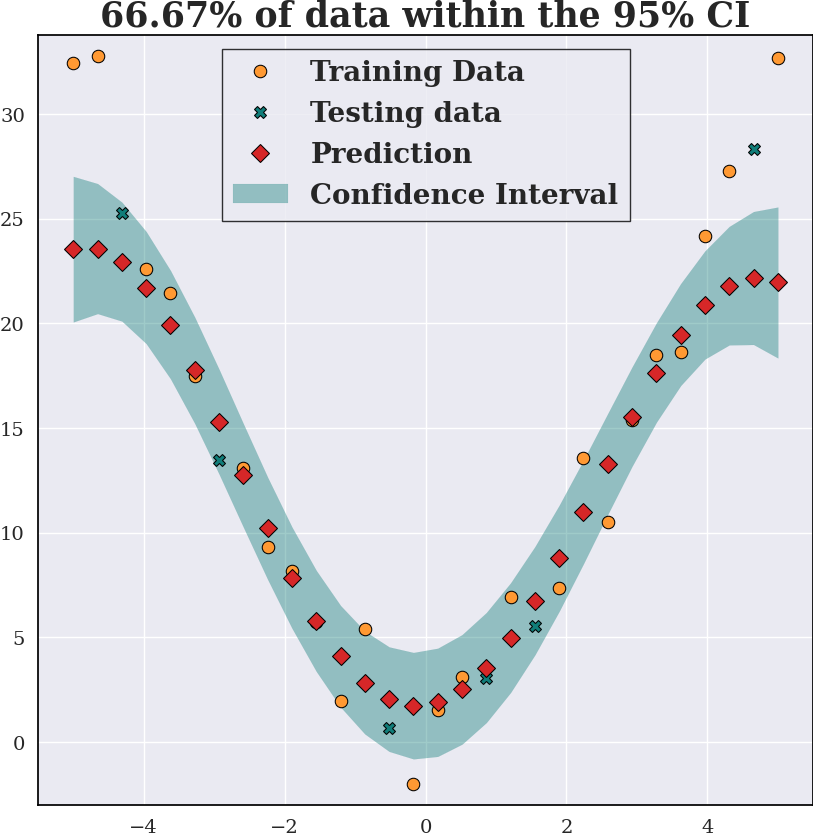}
    \includegraphics[width=.28\textwidth]{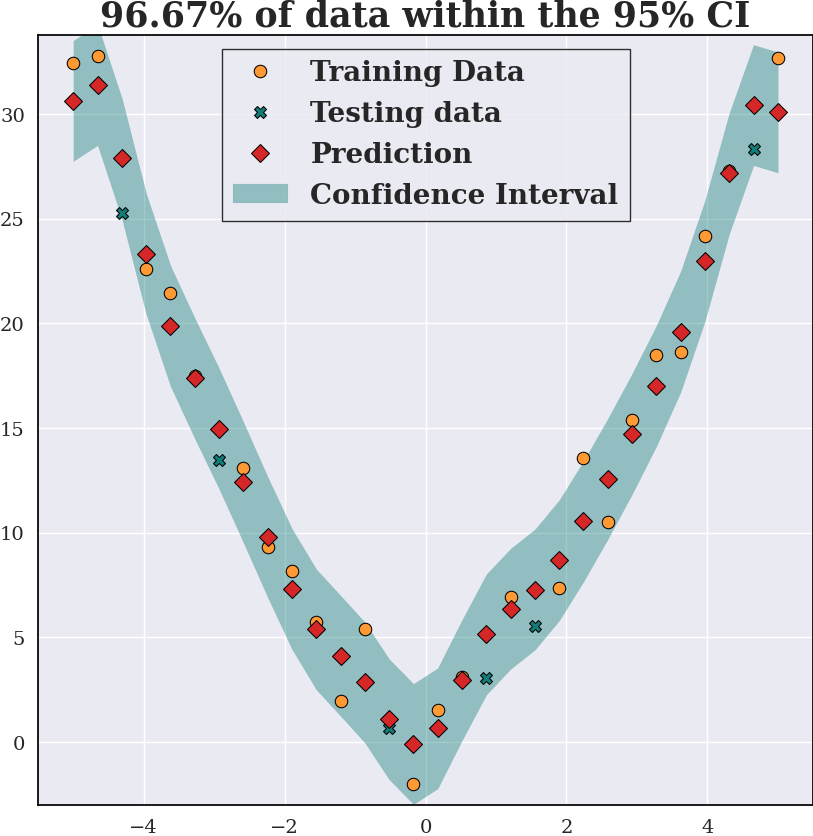}
    \caption{The left panel displays the observed targets \( y_i \) and true function values \( f(x_i) \) for 30 points \( x_i \) uniformly distributed in the interval \([-5, 5]\). The targets are generated by \( y_i = f(x_i) + \epsilon_i \), where \( f(x) = 3|x|^{\frac{3}{2}} + 2\sin(2\pi x) \) and \( \epsilon_i \sim N(0, 1) \). The middle panel illustrates predictions and 95\% confidence intervals from a zero-mean Gaussian Process (GP). The right panel presents results from the proposed two-stage GP approach (Algorithm \ref{alg:two-stage GP}). \textbf{The Exact-GP underfits the data, with its 95\% confidence interval covering only 66.7\% of the data, while the two-stage GP covers 96.67\% of the data.}
}
    \label{fig:proposition-1}
\end{figure}
\section{Supplementary Materials for Section \ref{sec: kernel misspecification}}
Before we proceed to the theoretical justification of this intuition, we introduce a few widely used assumptions \cite{wang2022gaussian}. These assumptions will be made throughout the paper for theoretical analysis.
\begin{itemize}
    \item \textbf{C1 (Interior Cone Condition)}: The domain of interest $\Omega \subset \mathbb{R}^{d}$ is a compact set with positive Lebesgue measure and Lipschtz boundary, and satisfies the interior cone condition, which means there exists an $\theta \in (0, 2\pi)$ and $R > 0$ such that $\forall x \in \Omega$, there exists a unit vector $\xi(x)$ such that the cone $\mathcal{C} = \{x + \lambda y, \lambda \in [0, R], \|y\| =1, \xi(x)^{\top}y \geq \cos{(\theta)} \}$ will be contained in $\Omega$. 
    \item \textbf{C2 (Integrability of the Ground-truth Kernel)}: There exits $m_0 > \frac{d}{2}$ such that,
    \begin{align*}
        C_1(1+\|\omega\|_2^2)^{-m_0} \leq \mathcal{F}(k^{*})(\omega) \leq C_2(1+\|\omega\|_2^2)^{-m_0}, \forall \omega \in \RR^{d},
    \end{align*}
    where $k^*$ is the ground-truth kernel function such that $f\in \mathcal{H}_{k^*}$, $\mathcal{F}$ denotes the Fourier transform, $C_1$ and $C_2$ are two universal constants.
    \item \textbf{C3 (Integrability of the Specified Kernel)}: There exits $m > \frac{d}{2}$ such that,
    \begin{align*}
        C_3(1+\|\omega\|_2^2)^{-m} \leq \mathcal{F}(k)(\omega) \leq C_4(1+\|\omega\|_2^2)^{-m}, \forall \omega \in \RR^{d},
    \end{align*}
    where $k$ is the specified kernel for GPR, it is likely misspecified such that $f\notin \mathcal{H}_{k}$.
    \item \textbf{C4 (Quasi-Uniform Design)}: Let $\mathcal{X} = \{X_1, X_2, X_3, \dots, \}$ be a sequence of designs with $|X_i| = i$. The fill distance of $X_n$ is defined by 
    $h_{X_{n}, \Omega} = \sup_{x\in \Omega}d(x, X_n)$, where $d(x, X_n)$ is the distance between the sampling points $X_n$ and $x$. The separation distance is defined as $q_{X_n} = \min_{x_i, x_j \in X_n}\frac{1}{2}\|x_i - x_j\|_2$.  Assume $h_{X_{n}, \Omega} \leq C n^{-\frac{1}{d}}$ and $q_{X_n}\geq C'n^{-\frac{1}{d}} $, where $C$ and $C'$ are two universal constants only depending on $\Omega$.
\end{itemize}
Assumptions C1-C4 are common assumptions in the KRR/GP literature, C1 ensures the regularity of the domain, C2-C3 ensure the integrability of Fourier transform of the ground-truth and specified kernels, and many kernels satisfy this condition including Mat\'ern kernel and RBF kernel. C4 can be achieved using the Farthest Point Sampling (FPS) method \cite{zhao2023adaptive} which is detailed in Algorithm \ref{alg:fps} in Appendix \ref{appendix: supplementary Materials for sec3}. With the specified assumptions in place, we can establish a probabilistic bound on the ratio of the prediction error to the irreducible error. This measures how effectively a well-specified model is likely to perform in comparison to the inherent noise in the data.
The following FPS algorithm can generate quasi-unifrom design with cost $O(nd)$. The quasi-uniform design in \textbf{Assumption C4} is not only important for theoretical analysis. Actually, it can enhance the structure in the covariance matrix of KRR/GPR for more efficient computation. For more details, please c.f. \cite{zhao2023adaptive}.
\label{appendix: supplementary Materials for sec3}
\begin{algorithm}[H]
\caption{Farthest Point Sampling (\texttt{FPS})}
\begin{algorithmic}[1]
\State \textbf{Input}: Dataset $X$ of size $n$, number of samples $k$
\State \textbf{Output}: Landmark point set $X_k$ of size $k$
\State Assign index to data points in $X$ as $\bx_1, \ldots, \bx_n$
\State Calculate $\bar{\bx} = \frac{1}{n} \sum_{i=1}^n \bx_i$
\State Set $l = \underset{1 \leq j \leq n}{\arg\min} \operatorname{dist}(\bx_j, \bar{\bx})$
\State Initialize the set $X_k = \{\bx_l\}$
\State Initialize the distance vector $\mathbf{d}$ of length $n$, setting all entries to $\infty$
\For{$i = 1$ to $k-1$}
    \State Update $\mathbf{d}$ with entries $\mathbf{d}(j) = \min\{\mathbf{d}(j), \operatorname{dist}(\bx_j, \bx_l)\}$
    \State Choose $l = \underset{1 \leq j \leq n}{\arg\max} \mathbf{d}(j)$
    \State Add $\bx_l$ to $X_k$
\EndFor
\State \textbf{Return}: $X_k$
\end{algorithmic}
\label{alg:fps}
\end{algorithm}

The first and most important misspecification needs to be considered is the misspecification of the functional form of the kernel function. If the hyperparameters are misspecified, there are many ways to recalibrate the hyperparameters to adjust the model such as \cite{Allison_Stephenson_F_Pyzer-Knapp_2023}. However, if the kernel is misspecified, then it's very hard to calibrate the model. The following theorem proved in \cite{fiedler2021practical} gives a probabilistic uniform error bound on the prediction error of GPR.
\begin{theorem}
\label{thm:frequentist-probabilistic-error bound}
For any $\delta \in (0, 1)$, and a function $f \in \mathcal{H}_{k}$ with $\|f\|_{k} \leq B$ for some $B\geq 0$ which generates the dataset $X_{*} = X \cup X^{*}$ where $X = \{\bx_i, y_i\}_{i=1}^{N}$ is the observed dataset and $X^{*} = {\bx_i^{*}}$ is the unobserved dataset. 
We choose a zero-mean GP prior $GP(\bo, k)$ to fit a GPR model on a random subset of the observed dataset of size $n$ and predict target for the unobserved locations $\bx_{*}$. Set $\bar{\sigma}_{\xi} = \max\{1, \sigma_{\xi}\}$ the following bound holds
    \begin{align}
        \PP[|\hat{m}_{n}(x) - f(x))| \leq \beta_{N}\hat{\sigma}_{n}(x), ~~ \forall n \in \NN,~\forall x \in X_{*}] \geq 1 - \delta
    \end{align}
where $\beta_n = B + \frac{1}{\sigma_{\xi}}\sqrt{\log{(|\frac{\bar{\sigma}_{\xi}}{\sigma_{\xi}}K_{XX} + \bar{\sigma}_{\xi}\bI_{n}|) - 2\log{\delta}}}$ and $\hat{\sigma}_{n}^2 =\hat{k}(x,x)$.
\end{theorem}
\begin{remark}
This bound gives an explicit probabilistic upper bound for the error between the ground-truth function and the predicted mean given by the GPR. If this bound is violated consistently, then we can draw the conclusion that $f$ is not in the RKHS of k. The model misspecification is indicated. This inspires us to propose the following algorithm to measure the model misspecification.
\end{remark}

\begin{algorithm}
\caption{Misspecification Checking Algorithm.}
\begin{algorithmic}[1] 
\State \textbf{Input}: Observed dataset $X$, kernel function $k$, a pre-defined desired probability $\delta$.
\State \textbf{Output}: Whether to reject hypothesis $H_0$: the model is well specified.
\For{$l=1$ \textbf{to} $100$}
    \State Randomly subsample a dataset of size 500 from $X$, and split it into $X_{\text{train}}$ and $X_{\text{test}}$.
    \State Training the GP model on the $X_{\text{train}}$ until NLL can not be decreased further.
    \State Compute the predicted means according to Equation \eqref{eq: GPR-predictions}.
    \State Compute $\Delta_i = |\hat{m}_{n}(\bx_i) - f(\bx_i)|$ for $\bx_i \in X_{\text{test}}$.
    \State Check if $\Delta_i \leq \beta_{n}\hat{\sigma}_{n}(\bx_i).$
    \State Compute empirical probability $\hat{P}_l = \frac{\#\{\Delta_i \leq \beta_{n}\hat{\sigma}_{n}(\bx_i)\}}{|X_{\text{test}}|}$.
\EndFor
\State Compute the mean $\hat{P}$ from the 100 $\hat{P}_{l}$s. 
\State \textbf{Return}: Reject the hypothesis if $\hat{P} < 1-\delta$; otherwise, accept the hypothesis.
\end{algorithmic}
\end{algorithm}
However, direct application of this algorithm is not feasible. Specifically, at line 7, we need to compute $|\hat{m}_{n}(\bx_i) - f(\bx_i)|$ for each $\bx_i \in X_{\text{test}}$, but $f(\bx_i)$ is unknown. Consequently, we compute $|\hat{m}_{n}(\bx_i) - y_i| = |\hat{m}_{n}(\bx_i) - (f(\bx_i) + \epsilon_i)|$ instead, and adjust line 8 to:
\begin{quote}
Check if $\Delta_i \leq \beta_{n}\hat{\sigma}_{n}(\bx_i) + F_{z}^{-1}(1-\frac{\delta}{2})\sigma_{\xi}$,
\end{quote}
where $F_{z}(x)$ is the CDF of the standard normal distribution. Another issue is the constant $B$ in $\beta_n$, which is problematic since, as mentioned by the authors in \cite{fiedler2021practical}, there are no reliable methods to estimate the RKHS norm for an arbitrary ground-truth function. To make this bound practically applicable, we propose eliminating the need to estimate the RKHS norm. The following lemma demonstrates that, under certain assumptions, the predictive posterior variance contracts towards zero as the size of the training data increases.

\begin{lemma}[Proposition 3 form \cite{wang2022gaussian}]
\label{lemma: contarction-rates}
Under assumptions C1-C4, if $\sigma_{\xi} > C n^{1-\frac{2m_0}{d}}$ for some constant $C$ there exists a constant C' independent of the size of $K_{XX}$ such that
\begin{align}
      \hat{\sigma}_{n}(x)^2 < C'(\sigma_{\xi}^2/n)^{1-\frac{d}{2m_0}}
\end{align}
\end{lemma}
Note that the additional assumption in the lemma, which requires $\sigma_{\xi} > C n^{1-\frac{2m_0}{d}}$, is not difficult to satisfy. Since $\sigma_{\xi}$ models the independent, irreducible noise, its value will typically be of $O(1)$ regardless of the training data size. However, as $n$ increases, $C n^{1-\frac{2m_0}{d}}$ decreases because $m_0 > \frac{d}{2}$. Additional lemmas are necessary to upper bound both the \textbf{Reducible Error} and the \textbf{Irreducible Error} of the predictions given by Gaussian Processes (GP).

This Lemma indicates that the \textbf{Reducible Error} can be bounded by the posterior variance.
\begin{lemma}[From Theorem 2 from \cite{chowdhury2017kernelized}]
\label{lemma: worst case error bound}
If $f \in \mathcal{H}_{k}$,
\begin{align}
    |K_{*X}(K_{{XX}} + \sigma^2_{\xi}\mathbf{I})^{-1}\mathbf{f}_{X}) - f(\bx_*)| \leq \|f\|_{\mathcal{H}_{k}}  \sigma_{\xi}
\end{align}  
where $g(\bx_i): = f(\bx_i) + \epsilon_i$ and $\mathcal{H}_{k^{\sigma}}$ is the RKHS for the kernel $k(\cdot,\cdot) + \sigma_{\xi}^2\delta(\cdot,\cdot)$ with $\delta(\bx,\by) = 1$ if $\bx = \by$ otherwise $\delta(\bx,\by) = 0$.
\end{lemma}


The following lemma provides a probabilistic lower bound for the \textbf{Irreducible Error}.
\begin{lemma}[\hyperlink{https://math.stackexchange.com/questions/4621017/lower-bound-for-the-gaussian-tail}{ref}]
 \label{lemma:tail lower bound for gaussian}
 \begin{align}
    \PP\big(~|K_{*X}(K_{{XX}} + \sigma^2_{\xi}\mathbf{I})^{-1}\mathbf{\epsilon_X}-\epsilon_{*}|) \geq A(\delta) \sigma~\big) \geq 1-\delta
    \end{align}
where $A(\delta) = \sqrt{1 - 2\log{(1-\delta)}} - 1$ and $\sigma = \sqrt{K_{*X}(K_{XX} + \sigma^2_{\xi}\mathbf{I})^{-2}K_{X*} + \sigma_{\epsilon}^{2}}$.
\end{lemma}
\begin{proof}
    For a standard normal random variable, its tail has the following lower bound.
    \begin{align}
    \PP\big(X \geq a) \geq e^{-\frac{a^2}{2} -a}
    \end{align}
    Since $\epsilon_i \sim N(0, \sigma_{\epsilon}^2)$,  $K_{*X}(K_{XX} + \sigma^2_{\xi}\mathbf{I})^{-1}\mathbf{\epsilon_X} -\epsilon_{*} \sim N(0, K_{*X}(K_{XX} + \sigma^2_{\xi}\mathbf{I})^{-2})K_{X*} + \sigma_{\epsilon}^{2})$.
    Therefore,set $\sigma = \sqrt{K_{*X}(K_XX + \sigma^2_{\xi}\mathbf{I})^{-2}K_{X*} + \sigma_{\epsilon}^{2}}$ we know
    \begin{align}
    \PP\big(\frac{K_{*X}(K_{{XX}} + \sigma^2_{\xi}\mathbf{I})^{-1}\mathbf{\epsilon_X} - \epsilon_{*}}{\sigma} \geq a) \geq e^{-\frac{\delta^2}{2} -a}
    \end{align}
    Set $A(\delta) = \sqrt{1 - 2\log{(1-\delta)}} - 1$, we derive
    \begin{align*}
    \PP\big(|K_{*X}(K_{{XX}} + \sigma^2_{\xi}\mathbf{I})^{-1}\mathbf{\epsilon_X} - \epsilon_{*}|) \geq A(\delta) \sigma) \geq 1-\delta
    \end{align*}
    Using the symmetry of the Gaussian random variable gives the desired result.
\end{proof}
Next lemma provides an explicit lower bound for the constant $\sigma$ in previous lemma.
\begin{lemma}
\label{lemma: lower bound for transformed Gaussian}
   We have a uniform lower bound for the variance $\sigma$ in  Lemma \ref{lemma:tail lower bound for gaussian},
   \begin{align*}
       \sigma_{*} \geq \frac{\sqrt{K_{*X}K_{X*}}}{\lambda_{1} + \sigma_{\xi}^2}
   \end{align*}
\end{lemma}
\begin{proof}
    Assume $\lambda_1 \geq \lambda_2\geq \dots\geq \lambda_n>0$ are the eigenvalues of $K_{XX}$. $(K_XX + \sigma^2_{\xi}\mathbf{I}) \prec (\lambda_1 + \sigma_{\xi}^2)\bI.$ $K_{*X}(K_{XX} + \sigma^2_{\xi}\mathbf{I})^{-2}K_{X*} \geq \frac{K_{*X}K_{X*}}{(\lambda_{1} + \sigma_{\xi}^2)^2}$ . Therefore, $ \sigma_{*} = \sqrt{K_{*X}(K_XX + \sigma^2_{\xi}\mathbf{I})^{-2}K_{X*} + \sigma_{\epsilon}^{2}} \geq \frac{\sqrt{K_{*X}K_{X*}}}{\lambda_{1} + \sigma_{\xi}^2}$.
\end{proof}
Combining all Lemma together, we can proceed to the proof of the Theorem \ref{thm:model-misspecification-bound}.
\begin{proof}[Proof of Theorem \ref{thm:model-misspecification-bound}]
\begin{align*}
    |\hat{m}_{n}(\bx_*) - (f(\bx_*)+\epsilon_{*})| & = |K_{*X}(K_{{XX}} + \sigma^2_{\xi}\mathbf{I})^{-1}(f_{X}+\mathbf{\epsilon_X}) -  (f(\bx_*)+\epsilon_{*})| \\
    & \leq |K_{*X}(K_{{XX}} + \sigma^2_{\xi}\mathbf{I})^{-1}f_{X} - f(\bx_*)| + |K_{*X}(K_{{XX}} + \sigma^2_{\xi}\mathbf{I})^{-1}\mathbf{\epsilon_X})  - \epsilon_{*}| \\
    & \leq \|f\|_{\mathcal{H}_{k}}  C'(\sigma_{\xi}^2/n)^{1-\frac{d}{2m_0}} + |K_{*X}(K_{{XX}} + \sigma^2_{\xi}\mathbf{I})^{-1}\mathbf{\epsilon_X} - \epsilon_{*}|
\end{align*}
According to Lemma \ref{lemma:tail lower bound for gaussian} and Lemma \ref{lemma: lower bound for transformed Gaussian}, we know
\begin{align*}
    \PP\big(~|K_{*X}(K_{{XX}} + \sigma^2_{\xi}\mathbf{I})^{-1}\mathbf{\epsilon_X}-\epsilon_{*}|  \geq A(\delta) \sigma_{*} ~\big) \geq 1-\delta,
    \end{align*} 
where $ \sigma_{*} \geq \frac{\sqrt{K_{*X}K_{X*}}}{\lambda_{1} + \sigma_{\xi}^2}$
Then we divide both sides by $|K_{*X}(K_{{XX}} + \sigma^2_{\xi}\mathbf{I})^{-1}\mathbf{\epsilon_X} - \epsilon_{*} |$,
With probability $1- \delta$, 
\begin{align*}
    \frac{|\hat{m}_{n}(X_i) - f(x)|}{|K_{*X}(K_{{XX}} + \sigma^2_{\xi}\mathbf{I})^{-1}\mathbf{\epsilon_X}) - \epsilon_{*}|} 
    & \leq 1 + \|f\|_{\mathcal{H}_{k}}C'\frac{(\sigma_{\xi}^2/n)^{1-\frac{d}{2m_0}}}{|K_{*X}(K_{{XX}} + \sigma^2_{\xi}\mathbf{I})^{-1}\mathbf{\epsilon_X} - \epsilon_{*} |} \\
    & \leq 1 + \frac{\|f\|_{\mathcal{H}_{k}}C'(\sigma_{\xi}^2/n)^{1-\frac{d}{2m_0}}}{A(\delta)\sigma_{*}}
\end{align*}
when $n \geq \frac{\sigma_{\xi}^2}{\Big(\frac{0.01A(\delta)\sqrt{K_{*X}K_{X*}}}{(\lambda_{1} + \sigma_{\xi}^2)C'\|f\|_{\mathcal{H}_{k}}}\Big)^{\frac{2m_0}{2m_0 - d}}} $,
we derive
\begin{align*}
    \frac{|\hat{m}_{n}(X_i) - f(x)|}{|K_{*X}(K_{{XX}} + \sigma^2_{\xi}\mathbf{I})^{-1}\mathbf{\epsilon_X} - \epsilon_{*}|} 
    & \leq 1.01.
\end{align*}
\end{proof}
\textbf{Experimental Details of Table \ref{tab:kernel checking metrics}:} In this ablation study, we we run Exact-GP with and without the AKS in Algorithm \ref{alg:automatic kernel searching} for seven small to medium scale UCI regression datasets. All the datasets are proprecessed as  \cite{gal2016dropout}~\footnote{The dataset can be downloaded from \hyperlink{https://github.com/yaringal/DropoutUncertaintyExps}.{https://github.com/yaringal/DropoutUncertaintyExps}}. We use an AdamW optimizer with learning rate $0.1$ and the learn rate scheduler get\_cosine\_with\_hard\_restarts\_schedule\_with\_warmup. We run the optimization for $100$ iterations, which can guarantee the NLL in Equation \eqref{eq:scaled-likelihood} has converged for all datasets.

\section{Supplementary Materials for Section \ref{sec: subsampling GP}}
\label{appendix: supplementary Materials for sec4}
For simplicity, we will ignore the additive constant and constant scaling factors in Equation \eqref{eq: GPR-predictions} and use
\begin{align}
    L(\theta;X_n) &= \frac{1}{n}(\by_{n}^{\top}K_{n}^{-1}\by + \log{\det{K_n}}) \\
    \frac{\partial L(\theta;X_n)}{\partial \theta} & =\frac{1}{n}
    \Tr{K_{n}^{-1}(\bI_{n} - \by_{n}\by_{n}^{\top}K_{n}^{-1})\frac{\partial K_n}{\partial \theta}} 
\end{align}
\label{sec: supplementary Materials for sec4}
The following convex concentration property is a very common assumption for proving matrix concentration inequality. 
\begin{definition}[Convex Concentration Property]
\label{def:convex concentartion property}
A random vector $X \in \mathbb{R}^{n}$ satisfies the convex concentration property with constant $M$ if for every $1-$Lipschitz convex function $\phi(\cdot):\mathbb{R}^{n} \rightarrow \mathbb{R} $, $\EE|\phi(X)|< \infty$ and the following concentration inequality holds
\begin{align*}
    \PP(|\phi(X) - \EE[\phi(X)]| \geq t) \leq 2e^{-\frac{t^2}{M^2}}
\end{align*}
\end{definition}
The following theorem shows that the Gaussian vector with i.i.d standard normal components satisfies the convex concentration property.
\begin{theorem}[Theorem 5.6 from \cite{boucheronconcentration}]
\label{thm: i.i.d gaussian tail}
 Let $X \in \RR^{d}$ be a vector of n independent standard normal random variables, then for a $L-$Lipschitz convex function $\phi(\cdot):\mathbb{R}^{n} \rightarrow \mathbb{R} $, the following bound holds
 \begin{align*}
     \PP(|\phi(X) - \EE[\phi(X)]| \geq t) \leq 2e^{-\frac{t^2}{2L^2}}
 \end{align*}
\end{theorem}
Hanson-Wright inequality is a concentration inequality for bounding the tail of a quadratic form. The following bound from \cite{adamczak2015note} provide an uniform tail bound which could be used for bound the difference between the quadratic from in NLL of GPR.
\begin{theorem}[Uniform Hanson-Wright Inequality (Theorem 2.3 of \cite{adamczak2015note})]
\label{thm:Hanson-Wright}
For any random vector $X \in \mathbb{R}^{n}$ with mean zero satisfying the convex concentration property with constant $K$ and any square matrix $A \in \mathbb{R}^{n\times n}$, the following concentration bound holds for any $t>0$ and some universal constant $C$
\begin{align*}
    \mathbb{P}(|X^{\top}AX - \mathbb{E}[X^{\top}AX]| \geq t) \leq 2\exp{\Big(-\frac{1}{C}\min{\Big\{\frac{t^2}{M^2\|A\|_{F}\|Cov(X)\|}, \frac{t}{M^2\|A\|}\Big\}}\Big)}
\end{align*}
\end{theorem}
With this Hanson-Wright Inequality, we can prove the following concentration inequality for the NLL.
\begin{theorem}
\label{thm: concentration of loss}
  The following concentration inequality hold,
\begin{align}
    \PP(|L({\theta;n}) - EL(\theta;n)| > t) \leq 2\exp{\Big(-\frac{1}{C}\min{\Big\{\frac{n^2t^2}{2\|\Tilde{K}\|_{F}}, \frac{nt}{2\|\Tilde{K}\|}\Big\}}\Big)}
\end{align}
where $\Tilde{K} = U_{*}^{\top}K_{n}^{-1}(\theta)U_{*}$, $ U_{*}U_{*}^{\top}: =  (Q_{*}\Lambda_{*}^{\frac{1}{2}})(Q_{*}\Lambda_{*}^{\frac{1}{2}})^{\top}= K_{*}$ and $Q_{*}\Lambda_{*} Q_{*}^{\top}$ is $K_{n}(\theta^*)$'s eignevalue decomposition with the diagonals of the diagonal matrix $\Lambda^{*}$  being eigenvalues and columns of $Q_{*}$ the othonormal eigenvectors. Especially, when $\theta = \theta^{*}$,
\begin{align}
    \PP(|L({\theta;n}) - EL(\theta;n)|> t) \leq 2\exp{\Big(-\frac{1}{2C}\min{(nt^2, nt)}\Big)}
\end{align}
\end{theorem}
\begin{proof}
    For simplicity in the proof, we will use $K$ and $K_{*}$ to denote the kernel matrix with a set of generic hyperparameters $\theta$ and optimal hyperparameters $\theta^*$, respectively.
    Firstly, notice that there is no randomness in the second term of the $L(\theta;n)$. To be more specific, 
    $$L({\theta;n}) - EL(\theta;n) = \by^{\top}K^{-1}\by 
    -\Tr{K^{-1}K_{*}}.$$ We only care about deriving a tail bound for the quadratic form.
    $K_n(\theta)$ is always a positive definite matrix, since we assume $\sigma_{\xi} > 0$. Therefore, we have the othonormal eigendecomposition 
    \begin{align*}
            K = Q\Lambda Q^{\top} &= (Q\Lambda^{\frac{1}{2}})(Q\Lambda^{\frac{1}{2}})^{\top}:=UU^{\top} \\
            K_{*} = Q_{*} \Lambda_{*} Q_{*}^{\top} &= (Q_{*}\Lambda_{*}^{\frac{1}{2}})(Q_{*}\Lambda_{*}^{\frac{1}{2}})^{\top}:=U_{*}U_{*}^{\top}
    \end{align*}
    Thus we can rewrite the loss as 
    \begin{align*}
        L({\theta;n}) - EL(\theta;n) &= \by^{\top}K^{-1}\by 
    -\Tr{K^{-1}K_{*}} \\
    & = (U_{*}^{-1}\by)^{\top}U_{*}^{\top}K^{-1}U_{*}(U_{*}^{-1}\by) - \Tr{K^{-1}K_{*}}
    \end{align*}
    The transformed labels are still Gaussian with zero mean and the new covariance matrix
    $$Cov(U_{*}^{-1}\by) = U_{*}^{-1}Cov(\by)U_{*}^{-\top} = U_{*}^{-1}K_{*}U_{*}^{-\top} = \bI.$$
    This indicates that the components of $U_{*}^{-1}\by$ are i.i.d standard normal random variable since they are joint normal. Therefore, according to the Theorem \ref{thm: i.i.d gaussian tail}, we know $\Tilde{\by} = 
 U_{*}^{-1}\by$ satisfies the convex concentration property defined in Definition \ref{def:convex concentartion property} with $M = \sqrt{2}$. Then the Hanson-Wright inequality in Theorem \ref{thm:Hanson-Wright} applies to $\Tilde{\by}$ and $\Tilde{K} = U_{*}^{\top}K^{-1}U_{*}$, 
    \begin{align*}
        P(|\Tilde{\by}^{\top}\Tilde{K}\Tilde{\by} - \Tr{K^{-1}K_{*}}| > nt) & \leq 2\exp{\Big(-\frac{1}{C}\min{\Big\{\frac{n^2t^2}{2\|\Tilde{K}\|_{F}}, \frac{nt}{2\|\Tilde{K}\|}\Big\}}\Big)}
    \end{align*}
Dividing the both sides of the inequality in the probability, we derive
     \begin{align*}
     P(|L({\theta;n}) - EL(\theta;n)| &>t) \\
        &\leq 2\exp{\Big(-\frac{1}{C}\min{\Big\{\frac{n^2t^2}{2\|\Tilde{K}\|_{F}}, \frac{nt}{2\|\Tilde{K}\|}\Big\}}\Big)}
    \end{align*}
    when $\theta = \theta^*$, $\Tilde{K} = \bI$. This gives us the desired concentration inequality.
\end{proof}

\begin{remark}
    This theorem indicates that the probability of large average discrepancy between the $L({\theta;n})$ and $EL(\theta;n)$ is small. Especially, at the optimal hyperparameters, the tail probability of relative discrepancy larger than $t$ is of order $O(e^{-\min{(nt^2, nt)}}).$ 
\end{remark}
\begin{corollary}
\label{corollary: loss 1-delta bound}
With probability $1-\delta$, the following bounds hold,
\begin{align}
 |L({\theta;n}) - EL(\theta;n)| < \max{\Big\{ \frac{\sqrt{2C\|\Tilde{K}\|_{F}\log{\frac{2}{\delta}}}}{n}, \frac{2C\|\Tilde{K}\|\log{\frac{2}{\delta}}}{n}
 \Big\}}
\end{align}
Especially when $\theta = \theta^*$,
\begin{align}
  \frac{|L({\theta;n}) - EL(\theta;n)|}{n} < \sqrt{\frac{2C\log{\frac{2}{\delta}}}{n}}
\end{align}
\end{corollary}
\begin{proof}[Proof of Corollary \ref{corollary: loss 1-delta bound}]
\begin{align}
  |L({\theta;n}) - EL(\theta;n)| <\frac{\sqrt{2C\|\Tilde{K}\|_{F}\log{\frac{2}{\delta}}}}{n}
\end{align}
when $\frac{n^2t^2}{2\|\Tilde{K}\|_{F}} < \frac{nt}{2\|\Tilde{K}\|},$ otherwise
\begin{align}
  |L({\theta;n}) - EL(\theta;n)| < \frac{2C\|\Tilde{K}\|\log{\frac{2}{\delta}}}{n}
\end{align}
Combining these two cases gives us the desired results.
\end{proof}
Similarly, we can apply the Hanson-Wright concentration inequality to the gradient of loss.
\begin{theorem}
\label{thm: concentartion of gradient of loss}
 The following concentration inequality hold,
 \begin{align*}
        P(| \frac{\partial L(\theta;n)}{\partial \theta} - \frac{\partial EL(\theta;n)}{\partial \theta}|> t) 
        \leq 2\exp{\Big(-\frac{1}{C}\min{\Big\{\frac{n^2t^2}{2\|\Bar{K}\|_{F}}, \frac{nt}{2\|\Bar{K}\|}\Big\}}\Big)}
    \end{align*}
where $\Bar{K} = U_{*}^{\top}K^{-1}\frac{\partial K}{\partial \theta}K^{-1}U_{*}$. Especially, when $\theta = \theta^*$, we have
\begin{align*}
        P(| \frac{\partial L(\theta;n)}{\partial \theta}|_{\theta=\theta^*}> t) 
        \leq 2\exp{\Big(-\frac{1}{C}\min{\Big\{\frac{n^2t^2}{2\|\Bar{K}\|_{F}}, \frac{nt}{2\|\Bar{K}\|}\Big\}}\Big)}
 \end{align*}
\end{theorem}
\begin{proof}[Proof of Theorem \ref{thm: concentartion of gradient of loss}]
    With the same notation in the proof of Theorem \ref{thm: concentration of loss}.
    Again, notice that there is no randomness in the second term of the $\frac{\partial L(\theta;n)}{\partial \theta}$. To be more specific, 
    $$\frac{\partial L(\theta;n)}{\partial \theta} - \frac{\partial EL(\theta;n)}{\partial \theta} = -\by^{\top}K^{-1}\frac{\partial K}{\partial \theta}K^{-1}\by 
    +\Tr{K^{-1}\frac{\partial K}{\partial \theta}K^{-1}K_{*}}.$$ 
    Thus we can rewrite the gradient of loss as 
    \begin{align*}
        \frac{\partial L(\theta;n)}{\partial \theta} - \frac{\partial EL(\theta;n)}{\partial \theta} &= -\by^{\top}K^{-1}\frac{\partial K}{\partial \theta}K^{-1}\by 
    +\Tr{K^{-1}\frac{\partial K}{\partial \theta}K^{-1}K_{*}} \\
    & = -(U_{*}^{-1}\by)^{\top}U_{*}^{\top}K^{-1}\frac{\partial K}{\partial \theta}K^{-1}U_{*}(U_{*}^{-1}\by) - \Tr{K^{-1}\frac{\partial K}{\partial \theta}K^{-1}K_{*}}
    \end{align*}
    The transformed labels are still Gaussian with zero mean and the new covariance matrix
    $$Cov(U_{*}^{-1}\by) = U_{*}^{-1}Cov(\by)U_{*}^{-\top} = U_{*}^{-1}K_{*}U_{*}^{-\top} = \bI.$$
    This indicates that the components of $U_{*}^{-1}\by$ are i.i.d standard normal random variable since they are joint normal. Therefore, according to the Theorem \ref{thm: i.i.d gaussian tail}, we know $\Tilde{\by} = 
 U_{*}^{-1}\by$ satisfies the convex concentration property defined in Definition \ref{def:convex concentartion property} with $M = \sqrt{2}$. Then the Hanson-Wright inequality in Theorem \ref{thm:Hanson-Wright} applies to $\Tilde{\by}$ and $\Tilde{K} = U_{*}^{\top}K^{-1}\frac{\partial K}{\partial \theta}K^{-1}U_{*}$, 
    \begin{align*}
        P(|\Tilde{\by}^{\top}\Tilde{K}\Tilde{\by} - \Tr{K^{-1}\frac{\partial K}{\partial \theta}K^{-1}K_{*}}| > nt) & \leq 2\exp{\Big(-\frac{1}{C}\min{\Big\{\frac{n^2t^2}{2\|\Tilde{K}\|_{F}}, \frac{nt}{2\|\Tilde{K}\|}\Big\}}\Big)}\\
        & \
    \end{align*}
   Finally, we derive
        \begin{align*}
        P(| \frac{\partial L(\theta;n)}{\partial \theta} - \frac{\partial EL(\theta;n)}{\partial \theta}| &> t) \\
        &\leq 2\exp{\Big(-\frac{1}{C}\min{\Big\{\frac{n^2t^2}{2\|\Tilde{K}\|_{F}}, \frac{nt}{2\|\Tilde{K}\|}\Big\}}\Big)}\\
    \end{align*}
   This gives us the desired concentration inequality.
\end{proof}

\begin{corollary}
With probability $1-\delta$, the following bounds hold,
\begin{align}
 | \frac{\partial L(\theta;n)}{\partial \theta} - \frac{\partial EL(\theta;n)}{\partial \theta}|< \max{\Big\{ \frac{\sqrt{2C\|\Bar{K}\|_{F}\log{\frac{2}{\delta}}}}{n}, \frac{2C\|\Bar{K}\|\log{\frac{2}{\delta}}}{n}
 \Big\}}.
\end{align}
\end{corollary}
\begin{remark}
  This theorem further indicates that the probability of large average discrepancy between the $ \frac{\partial L(\theta;n)}{\partial \theta}$ and $\frac{\partial EL(\theta;n)}{\partial \theta}$ is small for large $n$ if only $\|\Bar{K}\|_{F}$ is growing slower than $n$.
\end{remark}
We can also prove a probabilistic bound for the relative discrepancy of NLL and EL.
\begin{theorem}
\label{thm: concentration of loss without scaling}
  The following concentration inequality hold,
\begin{align}
    \PP(\frac{|L({\theta;n}) - EL(\theta;n)|}{\Tr{K_{n}(\theta)^{-1}K_{n}(\theta^{*}) }} > t) \leq 2\exp{\Big(-\frac{1}{C}\min{\Big\{\frac{\Tr{\Tilde{K}}t^2}{2}, \frac{t}{2}\sum_{i=1}^{n}\frac{\lambda_i}{\lambda_1}\Big\}}\Big)}
\end{align}
where $\Tilde{K} = U_{*}^{\top}K_{n}^{-1}(\theta)U_{*}$, $\lambda_1 \leq \lambda_2 \leq \dots \leq \lambda_n$ and are $K_{n}(\theta^*)$'s eigenvalues and othonormal eigenvectors. Especially, when $\theta = \theta^{*}$,
\begin{align}
    \PP(\frac{|L({\theta;n}) - EL(\theta;n)|}{n} > t) \leq 2\exp{\Big(-\frac{1}{2C}\min{(nt^2, nt)}\Big)}
\end{align}
\end{theorem}

\begin{proof}[Proof of Theorem \ref{thm: concentration of loss without scaling}]
Similarly as the Proof of Theorem \ref{thm: concentration of loss},
    the Hanson-Wright inequality in Theorem \ref{thm:Hanson-Wright} applies to $\Tilde{\by}$ and $\Tilde{K} = U_{*}^{\top}K^{-1}U_{*}$, 
    \begin{align*}
        P(|\Tilde{\by}^{\top}\Tilde{K}\Tilde{\by} - \Tr{K^{-1}K_{*}}| > t) & \leq 2\exp{\Big(-\frac{1}{C}\min{\Big\{\frac{n^2t^2}{2\|\Tilde{K}\|_{F}}, \frac{nt}{2\|\Tilde{K}\|}\Big\}}\Big)}\\
        & \
    \end{align*}
    scale $t$ by $\Tr{K^{-1}K_{*}}$, we derive
        \begin{align*}
        P(|\Tilde{\by}^{\top}\Tilde{K}\Tilde{\by} - \Tr{K^{-1}K_{*}}| &> \Tr{K^{-1}K_{*}}t) \\
        &\leq 2\exp{\Big(-\frac{1}{C}\min{\Big\{\frac{\Tr{K^{-1}K_{*}}^2t^2}{2\|\Tilde{K}\|_{F}}, \frac{t\Tr{K^{-1}K_{*}}}{2\|\Tilde{K}\|}\Big\}}\Big)}\\
        &= 2\exp{\Big(-\frac{1}{C}\min{\Big\{\frac{\Tr{\Tilde{K}}^2t^2}{2\|\Tilde{K}\|_{F}}, \frac{t\Tr{\Tilde{K}}}{2\|\Tilde{K}\|}\Big\}}\Big)}\\
    \end{align*}
    Note $\|\Tilde{K}\|_{F} = \sqrt{\Tr{\Tilde{K}^2}} \leq \sqrt{\Tr{\Tilde{K}}^2} = \Tr{\Tilde{K}}$ and denote the eigenvalues of $\Tr{\Tilde{K}}$ by $\lambda_1 \leq \lambda_2 \leq \dots \leq \lambda_n$, we have
     \begin{align*}
        P(|\Tilde{\by}^{\top}\Tilde{K}\Tilde{\by} - \Tr{K^{-1}K_{*}}| &> \Tr{K^{-1}K_{*}}t) \\
        &\leq 2\exp{\Big(-\frac{1}{C}\min{\Big\{\frac{\Tr{\Tilde{K}}t^2}{2}, \frac{t}{2}(1+\frac{\lambda_2}{\lambda_1} + \dots + \frac{\lambda_n}{\lambda_1})}\Big)}\\
    \end{align*}
    when $\theta = \theta^*$, $\Tilde{K} = \bI$. This gives us the desired concentration inequality.
\end{proof}

\begin{corollary}
With probability $1-\delta$, the following bounds hold
\begin{align}
  \frac{|L({\theta;n}) - EL(\theta;n)|}{\Tr{K_{n}(\theta)^{-1}K_{n}(\theta^{*}) }} < \sqrt{\frac{2C\log{\frac{2}{\delta}}}{\Tr{\Tilde{K}}}}
\end{align}
when ${\frac{\Tr{\Tilde{K}}t^2}{2}\leq  \frac{t}{2}\sum_{i=1}^{n}\frac{\lambda_i}{\lambda_1}},$ otherwise
\begin{align}
  \frac{|L({\theta;n}) - EL(\theta;n)|}{\Tr{K_{n}(\theta)^{-1}K_{n}(\theta^{*}) }} < \sqrt{\frac{2C\log{\frac{2}{\delta}}\|\Tilde{K}\|}{\Tr{\Tilde{K}}}}
\end{align}
In summary,
\begin{align}
  \frac{|L({\theta;n}) - EL(\theta;n)|}{\Tr{K_{n}(\theta)^{-1}K_{n}(\theta^{*}) }} < \max{\Big\{ \sqrt{\frac{2C\log{\frac{2}{\delta}}\|\Tilde{K}\|}{\Tr{\Tilde{K}}}}, \sqrt{\frac{2C\log{\frac{2}{\delta}}}{\Tr{\Tilde{K}}}} \Big\}}
\end{align}
Especially when $\theta = \theta^*$,
\begin{align}
  \frac{|L({\theta;n}) - EL(\theta;n)|}{n} < \sqrt{\frac{2\log{\frac{2}{\delta}}}{n}}
\end{align}
\end{corollary}
The following bounds the gradient of NLL at the ground-truth hyperparameters.
\begin{theorem}
\label{thm: concentartion of gradient of loss without scaling}
 The following concentration inequality hold,
 \begin{align*}
        P(| \frac{\partial L(\theta;n)}{\partial \theta} - \frac{\partial EL(\theta;n)}{\partial \theta}|> t) 
        \leq 2\exp{\Big(-\frac{1}{C}\min{\Big\{\frac{t^2}{2\|\Bar{K}\|_{F}}, \frac{t}{2\|\Bar{K}\|}\Big\}}\Big)}
    \end{align*}
where $\Bar{K} = U_{*}^{\top}K^{-1}\frac{\partial K}{\partial \theta}K^{-1}U_{*}$. Especially, when $\theta = \theta^*$, we have
\begin{align*}
        P(| \frac{\partial L(\theta;n)}{\partial \theta}|_{\theta=\theta^*}> t) 
        \leq 2\exp{\Big(-\frac{1}{C}\min{\Big\{\frac{t^2}{2\|\Bar{K}\|_{F}}, \frac{t}{2\|\Bar{K}\|}\Big\}}\Big)}
 \end{align*}
\end{theorem}
\begin{proof}[Proof of Theorem \ref{thm: concentartion of gradient of loss without scaling}]
    With the same notation in the proof of Theorem \ref{thm: concentartion of gradient of loss}.
 Then the Hanson-Wright inequality in Theorem \ref{thm:Hanson-Wright} applies to $\Tilde{\by}$ and $\Tilde{K} = U_{*}^{\top}K^{-1}\frac{\partial K}{\partial \theta}K^{-1}U_{*}$, 
    \begin{align*}
        P(|\Tilde{\by}^{\top}\Tilde{K}\Tilde{\by} - \Tr{K^{-1}\frac{\partial K}{\partial \theta}K^{-1}K_{*}}| > t) & \leq 2\exp{\Big(-\frac{1}{C}\min{\Big\{\frac{t^2}{2\|\Tilde{K}\|_{F}}, \frac{t}{2\|\Tilde{K}\|}\Big\}}\Big)}\\
        & \
    \end{align*}
   Finally, we derive
        \begin{align*}
        P(| \frac{\partial L(\theta;n)}{\partial \theta} - \frac{\partial EL(\theta;n)}{\partial \theta}| &> t) \\
        &\leq 2\exp{\Big(-\frac{1}{C}\min{\Big\{\frac{t^2}{2\|\Tilde{K}\|_{F}}, \frac{t}{2\|\Tilde{K}\|}\Big\}}\Big)}\\
    \end{align*}
   This gives us the desired concentration inequality.
\end{proof}
Again, the concentration inequality can be used to derive the following probabilistic bound.
\begin{corollary}
\label{corollary: gradient 1-delta bound}
With probability $1-\delta$, the following bounds hold
\begin{align}
  | \frac{\partial L(\theta;n)}{\partial \theta} - \frac{\partial EL(\theta;n)}{\partial \theta}| \leq \sqrt{2C\|\Bar{K}\|_{F}\log{\frac{2}{\delta}}}
\end{align}
when $\frac{t^2}{2\|\Bar{K}\|_{F}}< \frac{t}{2\|\Bar{K}\|}$  otherwise
\begin{align}
  | \frac{\partial L(\theta;n)}{\partial \theta} - \frac{\partial EL(\theta;n)}{\partial \theta}| \leq 2C\|\Bar{K}\|\log{\frac{2}{\delta}}
\end{align}
In summary,
\begin{align}
 | \frac{\partial L(\theta;n)}{\partial \theta} - \frac{\partial EL(\theta;n)}{\partial \theta}|< \max{\Big\{ \sqrt{2C\|\Bar{K}\|_{F}\log{\frac{2}{\delta}}}, 2C\|\Bar{K}\|\log{\frac{2}{\delta}}}\Big\}
\end{align}
Especially when $\theta = \theta^*$,
\begin{align}
  \frac{|L({\theta;n}) - EL(\theta;n)|}{n} < \sqrt{\frac{2\log{\frac{2}{\delta}}}{n}}
\end{align}
\end{corollary}
Now with all the auxiliary lemmas and theorems, we can prove the Theorem \ref{thm: subsampled GP} as follows.
\begin{proof}[Proof of Theorem 3]
    In order to prove this, we compute the derivative of $EL(\theta)$ w.r.t the hyperparameters, and we discuss its properties for lengthscale, outputscale and noise parameter separately. When we are discussing one parameter, the other two parameters will be fixed. Firstly, we discuss the outputscale and noise parameters. Under which situation, the computations can be greatly simplified. When lengthscale $l$ is fixed to be the optimal lengthscale $l^{*}$, $K_n = \sigma_f^2K(l^*) + \sigma_{\xi}^2\bI$ and $K_{n}^{*} = \sigma_f^{*2} K(l^*) + \sigma_{\xi}^{*2}\bI$. Furthermore, we split it into two cases:\\
    \textbf{Case 1:} Varying outputscale $\sigma_{f}$ with fixed noise $\sigma_{\xi}^{2}$:\\
    In this case, notice that $K$ and $K_{*}$ can be simultaneously diagonalized. Suppose $K(l^*) = P^{\top}\Lambda P$, where $P^{\top}P = \bI$ then we have
    \begin{align*}
        EL(\sigma_f) & = \Tr{(\sigma_f^{2}K(l^*) + \sigma_{\xi}^{*2}\bI)^{-1}(\sigma_f^{*2} K(l^*) + \sigma_{\xi}^{*2}\bI)} + \log{\det{(\sigma_f^2K(l^*) + \sigma_{\xi}^{*2}\bI})}\\
        & = \sum_{i=1}^{n}\frac{\sigma_{f}^{*2}\lambda_i^{*} + \sigma_{\xi}^{*2}}{\sigma_{f}^{2}\lambda^{*}_i + \sigma_{\xi}^{*2}} + \sum_{i=1}^{n}\log{(\sigma_{f}^2\lambda_i^{*} + \sigma_{\xi}^{*2})}
    \end{align*}
    Therefore, the first order derivative can be computed 
    \begin{align*}
    \frac{\partial EL(\sigma_f)}{\partial \sigma_f}
    & = -\sum_{i=1}^{n}\frac{2\sigma_{f}\lambda_i^{*}( \sigma_{f}^{*2}\lambda_i^{*} + \sigma_{\xi}^{*2})}{(\sigma_{f}^{2}\lambda^{*}_i + \sigma_{\xi}^{*2})^2} + \sum_{i=1}^{n}
    \frac{2\sigma_{f}\lambda_i^{*}}{\sigma_{f}^{2}\lambda_i^{*} + \sigma_{\xi}^{*2}}\\
    & = \sum_{i=1}^{n}\frac{\lambda_i^{*2}}{(\sigma_{f}^{2}\lambda^{*}_i + \sigma_{\xi}^{*2})^2} 2\sigma_{f} (\sigma_{f}^{2} - \sigma_{f}^{*2})
    \end{align*}
    and second order derivative easily follows
    \begin{align*}
    \frac{\partial^2 EL(\sigma_f)}{\partial \sigma_f^2}
    & =\sum_{i=1}^{n}\lambda_i^{*2} \frac{(6\sigma_f^2 - 2\sigma_f^{*2})(\sigma_{f}^{2}\lambda^{*}_i + \sigma_{\xi}^{*2}) - 8\sigma_{f}^2 \lambda^{*}_i(\sigma_{f}^{2} - \sigma_{f}^{*2})}{(\sigma_{f}^{2}\lambda^{*}_i + \sigma_{\xi}^{*2})^3}.
    \end{align*}
    Therefore, 
    \begin{align*}
        \frac{\partial EL(\sigma_f)}{\partial \sigma_f} |_{\sigma_f = \sigma_f^{*}} = 0, \quad  \frac{\partial^2 EL(\sigma_f)}{\partial \sigma_f^2}|_{\sigma_f = \sigma_f^{*}} >0, 
    \end{align*}
   which means $\sigma_f^{*}$ is a local minimum and if $\sigma_{f} \in [\sigma_f^{*} - \epsilon, \sigma_f^{*} + \epsilon]$. It is guaranteed to converge to $\sigma_f^{*}$.
   \\
    \textbf{Case 2:} Varying noise $\sigma_{\xi}$ with fixed noise $\sigma_{\xi}^{2}$:
    \\
    In this case, notice that $K_n$ and $K_n^{*}$ can be simultaneously diagonalized. Suppose $K(l^*) = P^{\top}\Lambda P$, where $P^{\top}P = \bI$ then we have
    \begin{align*}
        EL(\sigma_{\xi}) & = \Tr{(\sigma_f^{*2}K(l^*) + \sigma_{\xi}^{2}\bI)^{-1}(\sigma_f^{*2} K(l^*) + \sigma_{\xi}^{*2}\bI)} + \log{\det{(\sigma_f^{*2}K(l^*) + \sigma_{\xi}^{*2}\bI})}\\
        & = \sum_{i=1}^{n}\frac{\sigma_{f}^{*2}\lambda_i^{*} + \sigma_{\xi}^{*2}}{\sigma_{f}^{*2}\lambda^{*}_i + \sigma_{\xi}^{2}} + \sum_{i=1}^{n}\log{(\sigma_{f}^{*2}\lambda_i^{*} + \sigma_{\xi}^{2})}
    \end{align*}
    Therefore, the first order derivative can be computed 
    \begin{align*}
    \frac{\partial EL(\sigma_{\xi})}{\partial \sigma_{\xi}}
    & = -\sum_{i=1}^{n}\frac{2\sigma_{\xi}( \sigma_{f}^{*2}\lambda_i^{*} + \sigma_{\xi}^{*2})}{(\sigma_{f}^{2}\lambda^{*}_i + \sigma_{\xi}^{2})^2} + \sum_{i=1}^{n}
    \frac{2\sigma_{\xi}}{\sigma_{f}^{*2}\lambda_i^{*} + \sigma_{\xi}^{2}}\\
    & = \sum_{i=1}^{n}\frac{1}{(\sigma_{f}^{*2}\lambda^{*}_i + \sigma_{\xi}^{*2})^2} 2\sigma_{\xi} (\sigma_{\xi}^{2} - \sigma_{\xi}^{*2})
    \end{align*}
    and second order derivative easily follows
    \begin{align*}
    \frac{\partial^2 EL(\sigma_{\xi})}{\partial \sigma_{\xi}^2}
    & =\sum_{i=1}^{n} \frac{(6\sigma_{\xi}^2 - 2\sigma_{\xi}^{*2})(\sigma_{f}^{2}\lambda^{*}_i + \sigma_{\xi}^{*2}) - 8\sigma_{\xi}^2 (\sigma_{\xi}^{2} - \sigma_{\xi}^{*2})}{(\sigma_{f}^{2}\lambda^{*}_i + \sigma_{\xi}^{*2})^3}.
    \end{align*}
    Therefore, 
    \begin{align*}
        \frac{\partial EL(\sigma_{\xi})}{\partial \sigma_{\xi}} |_{\sigma_{\xi} = \sigma_{\xi}^{*}} = 0, \quad  \frac{\partial^2 EL(\sigma_{\xi})}{\partial \sigma_{\xi}^2}|_{\sigma_{\xi} = \sigma_{\xi}^{*}} >0, 
    \end{align*}
   which means $\sigma_{\xi}^{*}$ is a local minimum and if $\sigma_{\xi} \in [\sigma_{\xi}^{*} - \epsilon, \sigma_{\xi}^{*} + \epsilon]$. It is guaranteed to converge to $\sigma_{\xi}^{*}$.
   For lengthscale $l$, we assume $\sigma_{f}$ and $\sigma_{\xi}$ are fixed to be $\sigma_{f}^*$ and $\sigma_{\xi}^*$.
   \\
    \textbf{Case 3:} Varying lengthscale $l$ with fixed outputscale parameter $\sigma_{f}^{*}$ and fixed noise $\sigma_{\xi}^{*}$:
    \\
    Since we can not diagonalize the $K$ and $K_{*}$ at the same time, we directly compute the gradient and the curvature.
    \begin{align}
     \frac{\partial EL(l)}{\partial l} &= \Tr{K_{n}(l)^{-1}(\bI_{n} - K_{n}(l_{*}) K_{n}(l)^{-1})\frac{\partial K_n}{\partial l}}
    \end{align}
    \begin{align}
    \frac{\partial^2 EL(l)}{\partial l^2}  &= \Tr{\frac{\partial K_{n}(l)^{-1}}{\partial l}(\bI_{n} - K_{n}(l_{*}) K_{n}(l)^{-1})\frac{\partial K_n}{\partial l}}\\
    &- \Tr{K_{n}(l)^{-1}\frac{\partial K_{n}(l)^{-1}}{\partial l}K_{n}(l_{*})\frac{\partial K_n}{\partial l}} \\
    & \Tr{K_{n}(l)^{-1}(\bI_{n} - K_{n}(l_{*}) K_{n}(l)^{-1})\frac{\partial^2 K_n(l)}{\partial l^2}}
    \end{align}
    Therefore,
     \begin{align*}
        \frac{\partial EL(l)}{\partial l} |_{l = l^{*}} = \Tr{K_{n}(l^{*})^{-1}(\bI_{n} - K_{n}(l_{*}) K_{n}(l^{*})^{-1})\frac{\partial K_n}{\partial l}} = 0
    \end{align*}
    \begin{align*}
        \frac{\partial^2 EL(l)}{\partial l^2} |_{l = l^{*}} &=
    \Tr{K_{n}(l^{*})^{-1} K_{n}(l^{*})^{-1}\frac{\partial K_n}{\partial l}\frac{\partial K_n}{\partial l}} \\
    & \geq \lambda_{n}(K_n^{-2}(l^{*}))\Tr{\big(\frac{\partial K_n}{\partial l}\big)^2} > 0
    \end{align*}
    This completes our proof of first part.
    In order to prove that $\theta^{*}$ is an $\epsilon$-optimal solution for $L(\theta;n)$, notice the simple fact that
    \begin{align*}
    L(\hat{\theta}^{*};n)  + (EL(\hat{\theta}^{*};n) - L(\hat{\theta}^{*};n))\leq L(\theta;n) + (EL(\hat{\theta};n) - L(\hat{\theta};n))
    \end{align*}
    Therefore, with probability $1-\delta$
    \begin{align*}
    L(\theta^{*};n) &\leq L(\theta;n) + (EL(\hat{\theta};n) - L(\hat{\theta};n)) - (EL(\hat{\theta}^{*};n) - L(\hat{\theta}^{*};n))\\
    &\leq L(\theta;n)  + 2\max{\Big\{ \frac{\sqrt{2C\|\Tilde{K}\|_{F}\log{\frac{2}{\delta}}}}{n}, \frac{2C\|\Tilde{K}\|\log{\frac{2}{\delta}}}{n}
 \Big\}}
    \end{align*}\\    
    Denote $\epsilon = 2\max{\Big\{ \frac{\sqrt{2C\|\Tilde{K}\|_{F}\log{\frac{2}{\delta}}}}{n}, \frac{2C\|\Tilde{K}\|\log{\frac{2}{\delta}}}{n}
 \Big\}}$ we complete the proof of the second part.
\end{proof}
\begin{remark}
Though the C5 assumption may initially appear far-fetched, it effectively abstracts the typical application of GPR in practice.
Users often commence with a chosen kernel (e.g., Matérn) and a default mean function (e.g., zero mean), assuming these choices are apt. Subsequently, an optimization procedure—employing methods such as gradient descent or cross-validation—is utilized to estimate the hyperparameters, which then are used as the optimal hyperparameters for predictions.
In the previous sections, we have introduced a pragmatic approach to mitigate the mean misspecification via a two-stage approach and evaluate the adequacy of the chosen kernel and mean function. Thus, this assumption mirrors the real-world methodology applied in the utilization of GPR.
\end{remark}
\begin{remark}
   In our analysis, we assumed that two of the three parameters remain fixed while proving convexity for the third. This approach is necessary due to the complex dynamics of the NLL surface, which are difficult to quantify because of the sophisticated Hessian. If all parameters were allowed to vary simultaneously, the techniques employed here would not be applicable, as $K$ and $K_{*}$
  would not share the same eigenspace.
\end{remark}
\textbf{Experimental Details of Figure \ref{fig:UCI wine contour plot}:} We use the first fold of the UCI wine dataset used in the previous section to plot this contour plot. Again, AdamW optimizer with learning rate $0.1$ is used and learn rate scheudler is used as previous experiment. The optimization is run for 50 steps. The plot can be reporduced by the lossContour.py file in githup repo. \\
\textbf{Supplementary Experiments to Figure \ref{fig:UCI wine contour plot}}
\begin{figure}[!htp]
\centering
\includegraphics[width=.23\textwidth]{figures/train_test_loss_contour_Lengthscale_Noise_0.1.png}
\includegraphics[width=.23\textwidth]{figures/train_test_loss_contour_Lengthscale_Noise_0.5.png}
\includegraphics[width=.23\textwidth]
{figures/train_test_loss_contour_Lengthscale_Noise_0.8.png}
\includegraphics[width=.245\textwidth, height=2.10cm]
{figures/train_test_loss_contour_Lengthscale_Noise.png} 
\includegraphics[width=.23\textwidth]{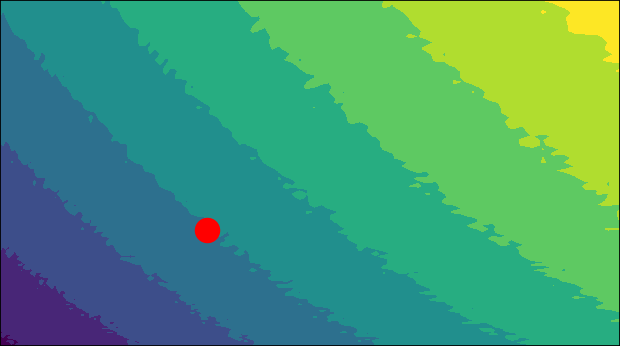}
\includegraphics[width=.23\textwidth]{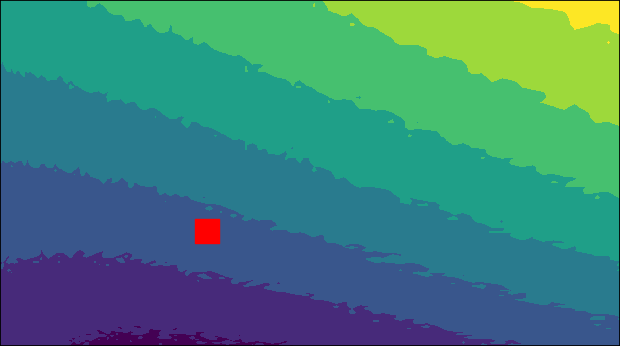}
\includegraphics[width=.23\textwidth]{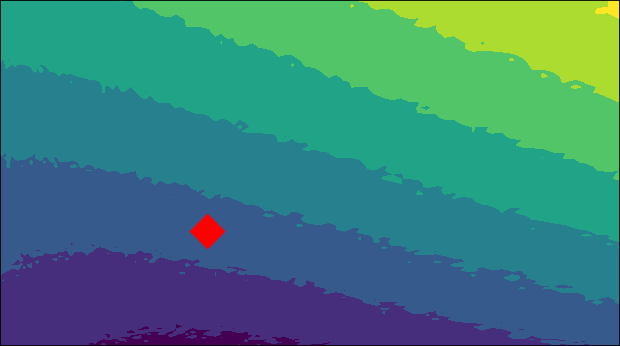}
\includegraphics[width=.245\textwidth]{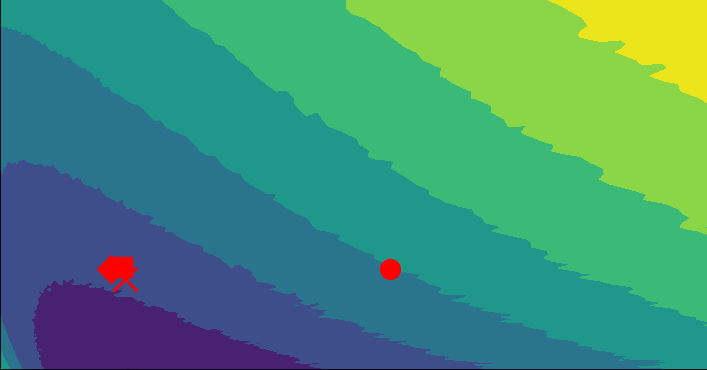}
\includegraphics[width=.23\textwidth]{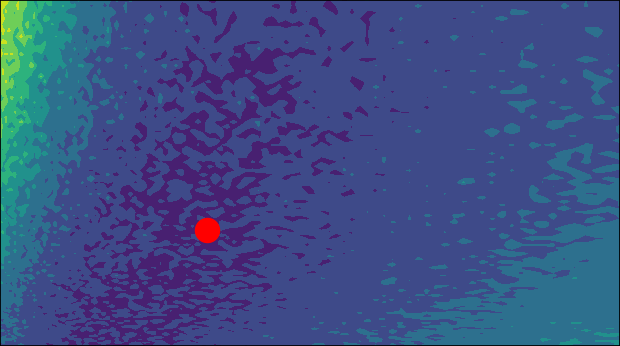}
\includegraphics[width=.23\textwidth]{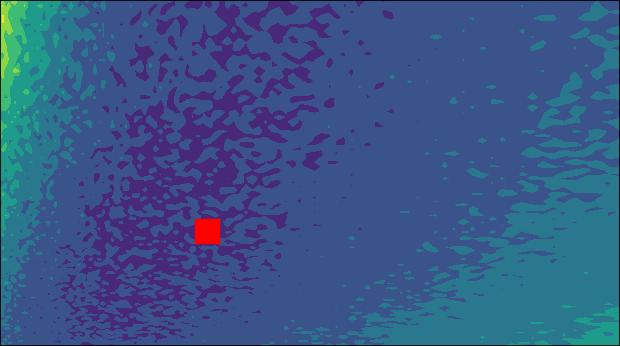}
\includegraphics[width=.23\textwidth]{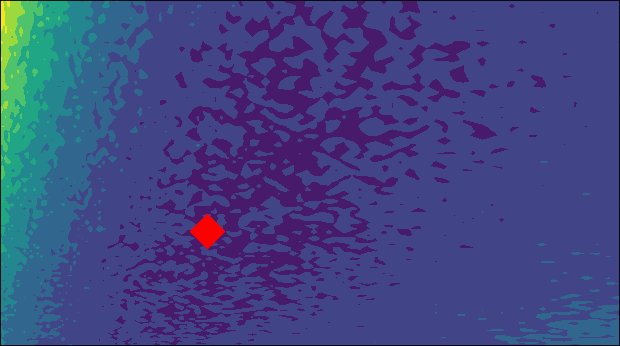}
\includegraphics[width=.245\textwidth]{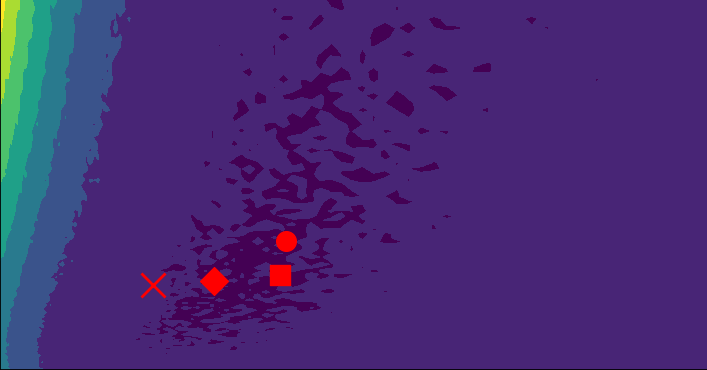}
    \caption{\textbf{Contour Plot of NLL for the UCI Wine Dataset}: This plot illustrates the pairwise contours of optimal lengthscale, outputscale, and noise. Each row represents contours between two parameters: lengthscale vs. noise, outputscale vs. noise, , and lengthscale vs. outputscale. Optimal hyperparameters for data subsets of 10\%, 50\%, 80\%, and 100\% are marked with red dot
    \begin{tikzpicture}
    \fill[red] (0,0) circle (3pt); 
    \end{tikzpicture}
    , square
    \begin{tikzpicture}
    \fill[red] (0,0) rectangle +(2mm,2mm);
    \end{tikzpicture}
    , diamond
    \protect\usebox{\diamondder}
    , and cross
    \protect\usebox{\cross}
    , respectively. Lighter areas indicate higher NLL values, while darker areas signify lower NLL values. All the contour plots are plotted on full training datasets.}
\label{fig:supplementary UCI wine contour plot}
\end{figure}
\subsection{Ablation Study on Exact-GP and SoD-Exact-GP}

In the following experiment, we evaluate the performance of two Gaussian Process (GP) methods: ExactGP and SoD-ExactGP, both employing the Mat\'ern-3/2 kernel. The ExactGP model is trained on the entire training dataset for 50 iterations using an Adam optimizer with a learning rate (\(lr\)) of $0.1$. In contrast, the SoD-ExactGP method involves a preliminary phase where only $200$ random samples from the dataset are used to train the model. This preliminary training also consists of $50$ iterations with an Adam optimizer, maintaining the same learning rate of $0.1$. The hyperparameters obtained from the SoD-ExactGP model serve as the initial values for an ExactGP model, which is then fine-tuned on the full training dataset for an additional $5$ iterations with an Adam optimizer, albeit at a reduced learning rate of $0.02$.
Table \ref{tab:exact_gp vs SoD exact_gp experimental_setup} summarizes the experimental setup for both the ExactGP and SoD-ExactGP methods, highlighting the key parameters and steps involved in each approach.
\begin{table}[ht]
\centering
\caption{Summary of Experimental Setup for ExactGP and SoD-ExactGP Methods}
\label{tab:exact_gp vs SoD exact_gp experimental_setup}
\begin{tabular}{l|c|c}
\hline
\textbf{Aspect} & \textbf{ExactGP} & \textbf{SoD-ExactGP} \\
\hline
Kernel & \multicolumn{2}{c}{Mat\'ern-3/2} \\
\hline
Training Samples & Full dataset & 200 random samples \\
\hline
Initial Iterations & 50 & 50 \\
\hline
Initial Learning Rate (\(lr\)) & 0.1 & 0.1 \\
\hline
Further Training & None & Full dataset for 5 iterations \\
\hline
Further Learning Rate (\(lr\)) & N/A & 0.02 \\
\hline
Optimizer & \multicolumn{2}{c}{Adam} \\
\hline
Training Samples for Prediction & Full dataset & Full dataset
\\
\hline
\end{tabular}
\end{table}

\begin{table}[h]
\centering
\caption{RMSE and NLL metrics for various datasets using Exact-GP with RBF kenrel and AKS-Exact-GP.}
\label{tab:Exact-GP vs SoD-Exact-GP metrics}
\renewcommand{\arraystretch}{1.2} 
\setlength{\tabcolsep}{3.0pt} 
\begin{tabular}{@{}lcccccccc@{}}
\toprule
\textbf{Dataset} & \multicolumn{3}{c}{\textbf{Exact-GP}} & \multicolumn{3}{c}{\textbf{SoD-Exact-GP}} \\
 \cmidrule(lr){2-4} \cmidrule(lr){5-7}
 & RMSE & NLL &QICE & RMSE & NLL &QICE \\ \midrule
Yacht   & $1.29\pm 0.51$& $-1.22\pm0.04$ & $5.00\pm0.83$&$1.26\pm 0.50$& $-1.25\pm0.03$ & $5.04\pm0.88$\\
Boston    & $2.68\pm 0.67$& $-0.55\pm0.09$ & $2.76\pm0.60$ & $2.67\pm 0.68$& $-0.57\pm0.12$ & $2.80\pm0.60$ \\
Energy   & $1.41\pm 0.21$& $-0.65\pm0.24$ & $2.82\pm0.67$ &  $1.49\pm 0.19$& $-0.67\pm0.06$ & $2.94\pm0.61$\\
Concrete & $5.17\pm 0.51$& $-0.27\pm0.21$ & $2.13\pm0.52$ & $5.30\pm 0.46$& $-0.10\pm0.16$ & $2.25\pm0.51$   \\
Wine    & $0.60\pm 0.05$& $0.31\pm0.60$ & $13.21\pm0.31$ &   $0.61\pm 0.04$& $0.90\pm0.12$ & $13.21\pm0.31$ \\
Kin8nm & $0.07\pm 0.00$& $-1.44\pm0.63$ & $0.83\pm0.19$   & $0.07\pm 0.00$& $-1.49\pm0.55$ & $0.80\pm0.19$ \\
Power  & $3.68\pm 0.22$& $2.7e5\pm1.2e6$ & $0.89\pm0.19$
& $3.81\pm 0.16$& $5.5e5\pm1.6e6$ & $0.96\pm0.17$\\
Naval  & $0.00\pm 0.00$& $1.6e3\pm5.6e3$ & $0.65\pm0.20$  &$0.00\pm 0.00$& $9.20\pm55.71$ & $0.65\pm0.19$ \\
\bottomrule
\end{tabular}
\end{table}
Table~\ref{tab:Exact-GP vs SoD-Exact-GP metrics} shows that training via subsampling warm start can usually achieve comparable performance similar to Exact-GP while it can reduce the training cost significantly.

\section{Supplementary Materials for Section \ref{sec: Two approaches for GP}}
In this section, we detail the complete algorithms for two recommended Gaussian Process (GP) approaches: the two-stage scalable GP and the two-stage Exact-GP. Our framework is quite flexible. For the two-stage scalable GP, any scalable GP approach can be used to replace the GPNN. For example, one can use SVGP, Vecchia Approximation, etc. Another flexibility in our framework is the ability to replace the KRR model with other predictive models, such as neural networks, ensemble models, etc.

\label{appendix: supplementary Materials for sec5}
\begin{algorithm}[!htp]
\caption{Scalable GP model}
\begin{algorithmic}[1] 
\State \textbf{Input}: Observed training dataset $X_{\text{train}} = \{\bx_i, y_i\}_{i=1}^{N_{\text{train}}}$, testing dataset $X_{\text{test}} = \{\Tilde{\bx}_i\}_{i=1}^{N_{\text{test}}}$, $m_{\text{train}}$ size of subsampled training set, $m_{\text{cal}}$ size of subsampled calibration set, $w$ number of nearest neighbors used for prediction.
\State \textbf{Output}: Mean prediction $\hat{m}(\cdot)$ and variance prediction $\hat{\sigma}^2(\cdot)$ for $X_{\text{test}}$ .
\State Call Algorithm \ref{alg:automatic kernel searching} to find the optimal kernel $k_{\text{mean}}$ for mean prediction.
\State Train a KRR model on a randomly subsampled size $m_{\text{train}}$ dataset of $X_{\text{train}}$ and and get the mean $\hat{m}$ using only w nearest neighbors.
\State Compute the demeaned training labels $y_i^{\circ} = y_i - \hat{m}(\bx_i)$
\State Call Algorithm \ref{alg:automatic kernel searching} to find the optimal kernel $k_{\text{var}}$ for variance prediction.
\State Train a zero-mean GP model $\mathcal{GP}(\bo, k_{\text{var}})$ on a randomly subsampled size dataset $X^{\text{train}}_{\text{train}}$ of $X_{\text{train}}^{\circ} = \{\bx_i, y_i^{\circ}\}_{i=1}^{N_{\text{train}}}$ and obtain the hyperparameters $\theta^{\square}_{var}$ and get the variance $\hat{\sigma}^2(\cdot)$.
\State \textbf{Return:} Mean prediction $\hat{m}(\cdot)$ and variance prediction $\hat{\sigma}^2(\cdot)$ for $X_{\text{test}}$.
\end{algorithmic}
\label{alg: Combined Scalable GP}
\end{algorithm}
\begin{algorithm}[!htp]
\caption{Exact GP model}
\begin{algorithmic}[1] 
\State \textbf{Input}: Observed training dataset $X_{\text{train}} = \{\bx_i, y_i\}_{i=1}^{N_{\text{train}}}$, testing dataset $X_{\text{test}} = \{\Tilde{\bx}_i\}_{i=1}^{N_{\text{test}}}$, $m_{\text{train}}$ size of subsampled training set.
\State \textbf{Output}: Mean prediction $\hat{m}(\cdot)$ and variance prediction $\hat{\sigma}^2(\cdot)$ for $X_{\text{test}}$ .
\State Call Algorithm \ref{alg:automatic kernel searching} to find the optimal kernel $k_{\text{mean}}$ for mean prediction.
\State Train KRR on full dataset $X_{\text{train}}$ with the $\theta^{\square}_{mean}$ as initialization and get the mean prediction $\hat{m}(\cdot)$.
\State Compute the demeaned training labels $y_i^{\circ} = y_i - \hat{m}(\bx_i)$.
\State Call Algorithm \ref{alg:automatic kernel searching} to find the optimal kernel $k_{\text{var}}$ for variance prediction on $X_{\text{train}}^{\circ} = \{\bx_i, y_i^{\circ}\}_{i=1}^{N_{\text{train}}}$.
\State Train a zero-mean GP model $\mathcal{GP}(\bo, k_{\text{var}})$ on a randomly subsampled size $m_{\text{train}}$ dataset of $X_{\text{train}}^{\circ}$ and and obtain the hyperparameters $\theta^{\square}_{var}$.
\State Train a zero-mean GP model $\mathcal{GP}(\bo, k_{\text{var}})$ on $X_{\text{train}}^{\circ}$ and get the variance prediction $\hat{\sigma}^2(\cdot)$.
\State \textbf{Return:} Mean prediction $\hat{m}(\cdot)$ and variance prediction $\hat{\sigma}^2(\cdot)$ for $X_{\text{test}}$.
\end{algorithmic}
\label{alg: Combined Exact GP}
\end{algorithm}

\section{Supplementary Materials for Section \ref{sec: numerical experiments}}
The proposed approach in this paper can be broadly applied to a wide range of downstream tasks, as uncertainty quantification is critical in many domains \cite{fan2024advanced,Datta_2022,fiedler2021practical,fan2024towards,Adlam_Lee_Padhy_Nado_Snoek_2023}. In section \ref{sec: numerical experiments}, we validate our methods through experiments on several real-world datasets, providing detailed results in the following section.
The first two experiments in Sec.\ref{sec: numerical experiments} are run on an Ubuntu 20.04.4 LTS machine equipped with 755
GB of system memory and a 24-core 3.0 GHz Intel Xeon Gold 6248R CPU. The last set of experiments involving pre-trained Foundation Models are run on the same machine with a H100 GPU.
\label{appendix: supplementary Materials for sec6}
\subsection{Experimental details of Sec.~\ref{sec: numerical experiements for exact-GP}}
All datasets used in this section are the same as those in Sec.~\ref{sec: kernel misspecification}. We use the AdamW optimizer for both Exact-GP and two-stage Exact GP. The initial learning rate is set to \(0.1\), and we use a learning rate scheduler, CosineAnnealingWarmRestarts, the same as in the previous section. The optimization runs for 100 iterations.

For the two-stage Exact GP, we use cross-validation to select the optimal regularization parameter and kernel for the KRR. Specifically, we search for the optimal regularization parameter from the set \(\{0.00001, 0.0001, 0.001, 0.01, 0.1, 1\}\) and the optimal kernel from either the RBF kernel or the Matern-1/2 kernel. We summarize all the key information to reproduce the experiments in Table \ref{tab:hyperparams_UQ_Exact-GP}.

\begin{table}[ht]
\centering
\caption{Summary of Implementation Details for ExactGP and Two Stage ExactGP in Sec.~\ref{sec: numerical experiements for exact-GP}.}
\label{tab:hyperparams_UQ_Exact-GP}
\begin{tabular}{l|c|c}
\hline
\textbf{Parameter} & \textbf{ExactGP} & \textbf{Two Stage Ex-GP} \\
\hline
First stage model & GP & KRR with cross validation\\ 
\hline
Second stage model &  N/A & AKS-GP\\ 
\hline
Training Samples & Full dataset & Subsampling training \\
\hline
Initial Iterations & 100 & 100 \\
\hline
Initial Learning Rate (\(lr\)) & 0.1 & 0.1 \\
\hline
Further Training & None & Full dataset for 10 iterations \\
\hline
Further Learning Rate (\(lr\)) & N/A & 0.1 \\
\hline
Optimizer & \multicolumn{2}{c}{AdamW} \\ 
\hline
Learning Rate Scheduler & \multicolumn{2}{c}{CosineAnnealingWarmRestarts}\\
\hline
\end{tabular}
\end{table}
 In the table, all the training-related parameters refer to the second stage GP model, as the hyperparameters of KRR are derived from cross-validation.

\subsection{Experimental details of Sec.~\ref{sec: UQ for GPNN}}
GPNN first trains the GPR on a randomly subsampled dataset, then calculates the miscalibration error 
\[
\text{cal} = \sum_{i=1}^{N_{\text{cal}}} \frac{(y_i - \hat{m}(\bx_i))^2}{\hat{\sigma}^2_i},
\]
where $N_{\text{cal}}$ is the size of the calibration dataset. If the prediction is perfect, this error should be close to 1. In reality, it may deviate significantly from 1.

To address this, \cite{Allison_Stephenson_F_Pyzer-Knapp_2023} proposed scaling the output scale parameter \( \sigma_f \) and noise parameter \( \sigma_{\xi} \) by this miscalibration error so that \( \text{cal} = 1 \) on the calibration dataset. We reproduce the GPNN results using the official code\footnote{GPNN official code: \url{https://github.com/ant-stephenson/gpnn-experiments/blob/main/experiments/method_of_paper.py}}. We use the same dataset and preprocessing methods as the GPNN paper. The data information is detailed in Table~\ref{tab:GPNN data}.
\begin{table}[ht]
    \centering
    \caption{Dataset Statistics}
    \begin{tabular}{lccccccc}
        \toprule
        & Poletele & Bike & Protein & Ctslice & Road3D & Song & HouseE \\
        \midrule
        $n$ & 4.6e+03 & 1.4e+04 & 3.6e+04 & 4.2e+04 & 3.4e+05 & 4.6e+05 & 1.6e+06 \\
        $d$ & 19 & 13 & 9 & 378 & 2 & 90 & 8 \\
        \bottomrule
    \end{tabular}
    \label{tab:GPNN data}
\end{table}
Since GPNN also used random subsampling training to obtain hyperparameters used for prediction, we adopt the same procedures. Our main differences are twofold: first, we do not include the calibration procedure. For each point, we first use a KRR with \( k \) nearest neighbors to predict the mean and then demean the entire training data. We pass the demeaned data to a GPNN model but only keep the variance prediction. Finally, the mean prediction is given by KRR prediction while the variance prediction is given by GPNN prediction. 
In contrast, GPNN first calibrates the hyperparameters, then uses the calibrated hyperparameters to predict the mean and variance for each point with a GP over its nearest neighbors. The second difference is that we select the kernel according AKS algorithm given in \ref{alg:automatic kernel searching}, while GPNN use a default kernel. We summarize all the key information in Table~\ref{tab:hyperparams_UQ_GPNN} for reproducing the results in Figure \ref{fig:uq_UCI}.
\begin{table}[ht]
\centering
\caption{Summary of Implementation Details for ExactGP and Two Stage ExactGP in Sec.~\ref{sec: UQ for GPNN}.}
\label{tab:hyperparams_UQ_GPNN}
\begin{tabular}{l|c|c}
\hline
\textbf{Parameter} & \textbf{GPNN} & \textbf{Two Stage GPNN} \\
\hline
First stage model & NA & KRRNN with cross validation\\ 
\hline
Second stage model &  GPNN & AKS-GPNN\\ 
\hline
\textbf{Number of Nearest Neighbors} & \multicolumn{2}{c}{50}\\
\hline
Iterations &\multicolumn{2}{c}{200}\\
\hline
Learning Rate (\(lr\)) &\multicolumn{2}{c}{0.1}\\
\hline
Optimizer & \multicolumn{2}{c}{AdamW} \\ 
\hline
Learning Rate Scheduler & \multicolumn{2}{c}{CosineAnnealingWarmRestarts}\\
\hline
Training Samples for Prediction & Full dataset & Full dataset \\
\hline
\end{tabular}
\end{table}

\subsection{Technical details of Sec.~\ref{sec: UQ for PFM}.}
The Gaussian process can not only work on regression problems but also on prediction problems. We study the potential utility of our proposed two-stage GP method for safety-critical clinical risk prediction tasks in Sec.~\ref{sec: UQ for PFM}. Specifically, we replace the default last fully-connected layer of the pre-trained foundation model with our GP classification layer, thus providing valuable uncertainty quantification to each test instance, by the stochastic prediction property brought by GP. 
We denote the process of representing a patient risk factor sample $\mathbf{x}$ into latent embedding $\mathbf{h}$ by
\begin{equation}
     \mathbf{h}=\phi_\omega(\mathbf{x}),
\end{equation}
where $\phi_\omega$ represents the PFM. A fully-connected layer that maps $h$ into the unnormalized classification output (\textit{i.e.} logits) by 
\begin{equation}
    \mathbf{z} = \mathbf{h}\mathbf{W}+\mathbf{b}.
\end{equation}
Here, $\mathbf{z} \in \mathbb{R}^{1\times C}$, where $C$ denotes number of target classes. The probabilities over the predicted classes can be obtained by 
\begin{equation}
    \Pr(y=c|x))=\argmax_c{ (\frac{e^{\mathbf{z}_{(c)}}}{\sum_{c'=1}^{C} e^{\mathbf{z}_{(c')}}})}.
\end{equation}
To adapt exact-GP for the classification problem, we treat predicting the logit $\mathbf{z}$ as a multi-output setting. Therefore, for each $\mathbf{z}_{(c')}$ that represents the logit for specific class $c'$, we train an exact-GP $\mathcal{GP}_{c'}(m_{c'}(\cdot), k_{c',\theta}(\cdot, \cdot))$ for it. In our implementation, we utilize the Dirichlet classification likelhood~\cite{milios2018dirichlet} and the soft one-hot encoding for further smooth optimization. Therefore, 
\begin{equation}
    \mathbf{z}_{c'} \sim \mathcal{GP}_{c'}(m_{c'}(\cdot), k_{c',\theta}(\cdot, \cdot)).
\end{equation}
For more details on the Gaussian process classification using the Dirichlet likelihood approach, we refer the reader to Appendix~\ref{appendix: background}.

For the implementation details, we provide Table~\ref{tab:hyperparams_UQ_PFM} for the ExactGP and our own Two Stage ExactGP. To replicate our experimental results, please refer to our codebase at this url\footnote{our codebase: \url{https://anonymous.4open.science/r/two-stage-GP-7906/GP-LLM-two-stage-GP/README.md}}. For the backbone PFMs, we use ``\textit{medicalai/ClinicalBERT}'' variant for ClinicalBERT with $768-d$ embedding, ``\textit{microsoft/biogpt}'' variant for BioGPT with $1024-d$ embedding, ``\textit{google/vit-base-patch16-224-in21k}'' variant for ViT with $768-d$ embedding. All PFM checkpoints can be downloaded from Huggingface Model Hub\footnote{Huggingface Model Hub: \url{https://huggingface.co/models}.}. During the fine-tuning, we freeze the learned parameters of PFMs, and only train the GP classification layer (one/two stage ExactGP) or the fully-connected layer (Monte Carlo Dropout).

\begin{table}[ht]
\centering
\caption{Summary of Implementation Details for ExactGP and Two Stage ExactGP in Sec.~\ref{sec: UQ for PFM}.}
\label{tab:hyperparams_UQ_PFM}
\begin{tabular}{l|c|c}
\hline
\textbf{ } & \textbf{ExactGP} & \textbf{Two Stage Ex-GP} \\
\hline
Kernel & \multicolumn{2}{c}{RBF} \\
\hline
Training Samples & Full dataset & 200 random samples \\
\hline
Initial Iterations & 600 & 50 \\
\hline
Initial Learning Rate (\(lr\)) & 0.1 & 0.1 \\
\hline
Further Training & None & Full dataset for 600 iterations \\
\hline
Further Learning Rate (\(lr\)) & N/A & 0.01 \\
\hline
Optimizer & \multicolumn{2}{c}{AdamW} \\ 
\hline
Learning Rate Scheduler & \multicolumn{2}{c}{CosineAnnealingWarmRestarts}\\
\hline
Training Samples for Prediction & Full dataset & Full dataset
\\
\hline
\end{tabular}
\end{table}

For the two datasets we used, we provide the statistics in Table~\ref{tab:data_stat}. For MedNLI, the input is a pair of sentences that consist of the medical history sentence serving as the \textit{premise} and the clinical outcome sentence serving as the \textit{hypotheses}. The prediction categories are \textit{entailment}, \textit{contradiction}, and \textit{neutral}. For BreakHis, the input is one breast cancer histopathological image, and the prediction targets include benign and malignant. Both datasets use patient risk factor data to predict potential clinical risk.

\begin{table}[h!]
    \centering
    \caption{Statistics of the used datasets.}
    \label{tab:data_stat}
    \begin{tabular}{c|c|c|c|cc}
    \toprule
    Dataset& Data Modality & Input Format& \#Category& \#Train & \#Test \\
    \midrule
    MedNLI~\cite{romanov2018mednli}& Text& a pair of sentences & 3 & 11,232 & 1,422 \\
    \hline
    BreakHis~\cite{spanhol2015breakhis}& Image& one RGB image& 2 & 5,005& 2,904\\
    \bottomrule
    \end{tabular}
    \vspace{-0.4cm}
\end{table}

\end{document}